\documentclass{article}

\usepackage{arxiv}

\usepackage[utf8]{inputenc}
\usepackage[T1]{fontenc}
\usepackage[pagebackref,breaklinks,colorlinks]{hyperref}
\usepackage{url}
\usepackage{booktabs}
\usepackage{amsfonts}
\usepackage{nicefrac}
\usepackage{microtype}
\usepackage{lipsum}

\usepackage{amssymb}
\usepackage{amsmath}
\usepackage[normalem]{ulem}
\usepackage{multirow}
\usepackage{array}
\usepackage{graphicx}
\usepackage{bm}
\usepackage{caption,subcaption}
\usepackage{pifont}
\usepackage{xcolor}
\usepackage{natbib}
\usepackage{tabularx}
\usepackage{makecell}

\title{GradAttn: Replacing Fixed Residual Connections with Task-Modulated Attention Pathways}

\author{
    Soudeep Ghoshal \\
    Kalinga Institute of Industrial Technology\\
    Bhubaneswar, India \\
    \texttt{2205421@kiit.ac.in} \\
    \And
    Himanshu Buckchash \\
    IMC University of Applied Sciences Krems\\
    Krems, Austria \\
    \texttt{himanshu.buckchash@imc.ac.at} \\
}

\begin{document}
\maketitle

\begin{abstract}
Deep ConvNets suffer from gradient signal degradation as network depth increases, limiting effective feature learning in complex architectures. ResNet addressed this through residual connections, but these fixed short circuits cannot adapt to varying input complexity or selectively emphasize task-relevant features across network hierarchies. This study introduces GradAttn, a variation of the residual approach in CNNs that replaces the fixed residual connections with attention-controlled gradient flow. By extracting multi-scale CNN features at different depths and regulating them through self-attention, \textit{GradAttn} dynamically weights shallow texture features and deep semantic representations. For representational analysis, we evaluated three \textit{GradAttn} variants across eight diverse datasets: from natural images and medical imaging to fashion recognition. The results demonstrate that \textit{GradAttn} outperforms ResNet-18 on five of eight datasets, achieving up to $+11.07\%$ accuracy improvement on FashionMNIST while maintaining a comparable network size. Gradient flow analysis reveals that controlled instabilities, introduced by attention, often coincide with improved generalization, challenging the assumption that perfect stability is optimal. Furthermore, positional encoding's effectiveness turned out to be dataset-dependent, with CNN hierarchies frequently encoding sufficient spatial structure. These findings render attention mechanisms as enablers of learnable gradient control, offering a new way for adaptive representation learning in deep neural architectures.
\end{abstract}


\section{Introduction} \label{section:introduction}

One of the main issues that has limited the depth of convolutional neural networks (CNNs) is the vanishing gradient problem, a situation where error signals diminish as they are propagated backward through multiple layers. The revolutionary introduction of residual connections in ResNet brought a major breakthrough in deep learning by preserving gradient flow, thereby enabling the successful training of very deep networks~\citep{he2016deep}. The prevalence of ResNet in academic research and its integration into various systems and applications suggest that it serves as a benchmark architecture due to its reliable performance and widespread acceptance~\citep{sima2024adaptive}. The core residual formulation, \(y = F(x, \{W_i\}) + x\), not only establishes gradient highways but also ensures that gradients are transmitted effectively across layers, mitigating errors during backpropagation.

Still, residual connections are inherently \textbf{static and uniform}: Each skip connection offers the same direct gradient paths, regardless of input complexity or the feature relevance of the particular task. Such uniformity neglects a fundamental characteristic: Different tasks require adaptive weighting of feature representations at various network depths. For complicated visual recognition tasks covering normal images to medical imaging, the needed learning requires the \textbf{adaptive weighting} of features at different depths of the network. It is not possible for residual connections to selectively highlight some routes while disabling others; they combine shallow texture features and deep semantic representations with the same level of importance.
Despite recent advances in hybrid architectures and attention mechanisms, only a few works treat gradient flow itself as a learnable, task-adaptive component. Most approaches either solve gradient degradation through fixed structural shortcuts or apply attention solely for feature enhancement, leaving a fundamental gap: the inability to dynamically route gradients based on input complexity and task requirements across network hierarchies.

This work, specifically \textbf{\textit{GradAttn}} (\href{https://github.com/SoudeepGhoshal/GradAttn}{Source Code}), introduces \textit{attention-controlled gradient flow}, replacing static residual connections with learnable gradient pathways. The proposed method, GradAttn, uses the attention mechanisms of the transformer to not only open up the features' interactions but also to enable gradient flow through CNN hierarchies. The major highlight of the method is the extraction of multi-scale CNN features from various depths and the mapping of these features to a common embedding space, where their contributions can be weighted dynamically by transformer attention; this thus creates gradient pathways that can be selected for holding global semantic patterns for complex scenes or for local texture cues of specialized domains, all completely learned automatically during the training process.

Our Contribution:
(a) We propose attention-controlled (inter-layer) gradient flow as a learnable alternative to fixed residual connections, enabling adaptive gradient pathways in deep networks.  
(b) We demonstrate that this approach outperforms ResNet-18 on five of eight diverse datasets, achieving up to $+11.07\%$ accuracy improvement while maintaining comparable parameter counts.  
(c) We provide empirical evidence that controlled gradient instabilities introduced by attention often coincide with improved generalization, challenging the assumption that perfect stability is always optimal.

We perform a direct comparison against the industry standard ResNet, confirming that gradient routing achieves more effective learning. Since CNNs are omnipresent, this work has broad and immediate applicability.

We investigate three architectural variants, viz., those without positional encoding (No PE), those with learnable positional encoding, and those with rotary positional encoding (RoPE), revealing that the effectiveness of explicit spatial encoding is dataset-dependent, with CNN hierarchies often providing sufficient structural priors. These findings position attention mechanisms as fundamental enablers of learnable gradient control, opening possibilities for adaptive deep learning across diverse visual domains.

The remainder of this paper is organized as follows. Section~\ref{section:related_works} reviews related work on gradient flow in deep architectures, attention mechanisms, and multi-scale feature integration. Section~\ref{section:method} details the proposed GradAttn architecture, its positional encoding variants, training protocols, and evaluation datasets. Section~\ref{section:results} presents comprehensive experimental results, including classification performance, domain-specific analysis, training dynamics, gradient flow analysis, and parameter efficiency. Finally, Section~\ref{section:conclusion} concludes this paper and outlines directions for future work.

\section{Related Works} \label{section:related_works}

\subsection{Gradient Flow in Deep Architectures}

Over the past few years, gradient flow innovations implemented as residual architectures have really changed the face of deep learning, especially in the area of convolutional neural networks (CNNs). ResNets (residual networks) rely on skip connections that allow gradients to travel around certain layers and thus empower the training of deeper networks~\citep{he2016deep}. Moreover, DenseNets link each layer with every other layer, thus encouraging feature reuse and maintaining gradient propagation throughout the network~\citep{huang2017densely}. EfficientNets advance these concepts through compound scaling methodologies that systematically balance network depth, width, and resolution, creating more sophisticated gradient flow patterns while maintaining computational efficiency~\citep{tan2019efficientnet}. Nevertheless, these architectures have a common feature, which is the dependence on fixed, uniform connections during training that may limit the flexibility of  gradient flow and adaptation to different input characteristics~\citep{he2016deep}. New changes in the field try to fix these problems; for example, as shown in~\citep{jastrzkebski2017residual}, residual connections support iterative inference, but they still function within the constraints of current architectures. It implies a demand for more versatile procedures that effectively allocate gradient flow depending on the fundamental requirements of the learning task.

Recent architectural innovations have further explored gradient flow optimization through structural modifications. ResNeXt extended ResNet's design by introducing cardinality as an additional dimension beyond depth and width, demonstrating that aggregating transformations can improve gradient propagation while maintaining computational efficiency~\citep{xie2017aggregated}. Inception architectures explored multi-branch convolutions that process information at different scales simultaneously, though these branches converge through concatenation rather than adaptive weighting~\citep{szegedy2015going}. More recently, Neural architecture search (NAS) has automated the discovery of optimal skip connection patterns, though these approaches still operate within the paradigm of fixed architectural decisions once deployment occurs~\citep{zoph2016neural}.

The degradation problem in deep networks has also been addressed through normalization techniques that stabilize gradient flow. Batch normalization reduces internal covariate shifts and enables higher learning rates, indirectly improving gradient propagation~\citep{ioffe2015batch}. Layer normalization and group normalization have extended these concepts to scenarios where batch statistics are unreliable, further demonstrating that gradient health depends on multiple architectural factors beyond skip connections alone~\citep{ba2016layer, wu2018group}. However, normalization techniques complement but do not replace the need for effective gradient routing mechanisms in very deep architectures.

More recent work has examined the instability introduced when learned residual mappings are left unconstrained. Xie et al.~\cite{xie2025mhc} proposed Manifold-Constrained Hyper-Connections (mHC), which extends the Hyper-Connections framework by projecting residual mapping matrices onto the Birkhoff polytope of doubly stochastic matrices via the Sinkhorn-Knopp algorithm. This constraint restores the identity mapping property across arbitrary network depths, preventing unbounded signal amplification. Empirical analyses of 27~B parameter language models showed that unconstrained composite residual mappings can reach gain magnitudes of approximately 3000, causing training instability at scale. GradAttn takes a complementary approach: rather than constraining the mapping space geometrically, the softmax normalization inherent to self-attention naturally bounds feature contributions, implicitly regularizing gradient routing without requiring explicit manifold projection.

\subsection{Attention Mechanisms and Hybrid Architectures}

Attention mechanisms have been the trend in NLP but have been integrated in CNNs to improve feature representation. On the one hand, works like Squeeze-and-Excitation Networks (SENets) and the Convolutional Block Attention Module (CBAM) have led to substantial advancements in image recognition fields by opening channel features \mbox{dynamically~\citep{hu2018squeeze, woo2018cbam}}. Besides this, the rise of Vision Transformers (ViTs) has brought along a new paradigm of hybrid models that combine CNNs and transformer architectures to make use of the advantages of both fields~\citep{dosovitskiy2020image}. Nevertheless, the majority of these traditional frameworks are quite limited, as they are mainly focused on attention for feature enhancement. Thus, they do not handle gradient flow efficiently, and hence, there is a gap between the methods that control (inter-layer) gradients during training remains.

Building on these foundations, the integration of attention mechanisms with convolutional architectures has evolved beyond simple feature recalibration. Non-local neural networks introduced self-attention blocks within CNNs to capture long-range dependencies, demonstrating that global context modeling enhances feature representations in vision tasks~\citep{wang2018non}. BAM (Bottleneck Attention Module) explored the dual-pathway design of attention across both spatial and channel dimensions, showing complementary benefits when applied at intermediate network stages~\citep{park2018bam}. More recently, CoAtNet architectures have systematically studied the interplay between convolutional inductive biases and transformer attention, revealing that hybrid designs can outperform pure CNN or pure transformer architectures when properly configured~\citep{dai2021coatnet}. Gradient flow in transformer architectures presents distinct challenges from CNNs.

In a related vein, InternImage~\citep{wang2023internimage} is a large-scale CNN foundation model that replaces fixed convolutional kernels with deformable convolutions as the core operator, enabling adaptive spatial aggregation conditioned on input content. Evaluated on ImageNet, COCO, and ADE20K, InternImage demonstrated that relaxing the rigid spatial inductive bias of standard convolutions yields strong gains, with InternImage-B achieving 84.9\% Top-1~accuracy on ImageNet-1K. However, InternImage's adaptivity operates within individual layers by dynamically adjusting sampling locations in the spatial domain, leaving the inter-layer flow of information governed by fixed residual connections.

Pre-normalization configurations have been shown to stabilize training in very deep models by preventing gradient explosion in attention layers~\citep{xiong2020layer}, while post-normalization variants demonstrate trade-offs between training stability and representational \mbox{capacity~\citep{touvron2021training}}.

Despite these advances, existing works treat gradient flow as either a stability problem to be solved through normalization and careful initialization or as a fixed architectural property through skip connections. None of these approaches consider gradient flow as a learnable, task-adaptive component of the network.

\subsection{Multi-Scale Feature Integration}

Multi-scale feature fusion strategies in detection and segmentation architectures relate to hierarchical feature integration. Feature pyramid networks (FPNs) combine features across multiple scales through lateral connections, establishing the importance of hierarchical feature integration~\citep{lin2017feature}. PANet further enhanced this design with bottom-up path augmentation, showing that bidirectional feature flow improves representation \mbox{learning~\citep{liu2018path}}. U-Net architectures demonstrate the effectiveness of skip connections that bridge encoder and decoder pathways in dense prediction tasks~\citep{ronneberger2015u}. However, these architectures employ fixed fusion patterns with predetermined connection topologies that do not adapt to input characteristics or task requirements.

Ensemble and multi-branch architectures provide another perspective on feature combination. Wide ResNets increased network capacity through wider layers rather than deeper stacks, showing that width provides complementary benefits to depth~\citep{zagoruyko2016wide}. Multi-Scale Dense Networks (MSDNets) combined features from different depths to enable early-exit inference, though feature combination weights remain static across all inputs~\citep{huang2017multi}. These methods demonstrate the value of combining information from multiple network locations but rely on fixed combination strategies that cannot selectively emphasize or de-emphasize specific feature hierarchies based on input complexity.

The attention-controlled gradient flow that we are suggesting is different from the fixed kinds of operations, as it gives a flexible design for the optimization of  gradient flow through which deep learning models can be made to work at high efficiency beyond the extent of residual and attention systems. In the next section, we describe the design of the proposed~method.

\section{Methods} \label{section:method}

The key idea of this work is to regulate the learning and flow of gradients via attention at the inter-layer level. Since each layer learns features at different levels of the \textit{semantic hierarchy}, the idea is to efficiently regulate  learning within these hierarchies. In this paper, we realize this key idea through self-attention in the image recognition problem using deep convolutional neural networks. The methodology is split into two parts: first, a high-level mathematical view of the proposed solution and, second, a deeper architectural~explanation.

\subsection{Conceptual Framework for Gradient Modulation}

We first establish a conceptual framework to reason about our multi‑scale token attention, which is also referenced as gradient modulation; the actual realization in GradAttn is through transformer self-attention, as described in Section~\ref{subsec:arch design}. Let a deep network be a composition of $L$ layers $\{\phi_l\}_{l=1}^{L}$ with parameters $\{\theta_l\}_{l=1}^{L}$, producing intermediate feature maps $f_l = \phi_l(f_{l-1}; \theta_l)$ for input $f_0 = x$. In a standard residual network, the backward gradient at layer $l$ is the unmodulated chain $g_l = \partial \mathcal{L} / \partial \theta_l$, and skip connections inject a uniform identity term $\partial f_{l+1} / \partial f_l = I + \partial F_l / \partial f_l$, treating every layer as equally update-worthy regardless of the input or training state. We argue that this is suboptimal: at any training step $t$, the useful update magnitude for layer $l$ depends on how much that layer has already learned relative to others, a quantity not visible from local gradients alone.

We therefore introduce a meta-regulator $\mathcal{M}_\psi$ with parameter $\psi$ that observes a global summary of the network state and emits a per-layer modulation signal $\alpha_l \in [0, 1]$, interpreted as the relative degree to which layer $l$ should be updated at step $t$. Conceptually, $\alpha_l$ encodes the notion that not all layers require equal adjustment at every training step: Some layers may have already converged to useful representations, while others remain under-adapted. It is crucial to distinguish between this analytical framework and our physical implementation. In our actual architecture, this modulation is achieved implicitly through transformer self-attention over the token sequence $Z = [z_1, \dots, z_L]$ from the extraction points defined in Equation~(\ref{eq:projection}). Specifically, the softmax operation within the attention layers dynamically weights the feature tokens. During backpropagation, these learned attention weights directly and naturally scale the magnitude of the gradient flow routed back to each corresponding convolutional stage. Therefore, the self-attention mechanism effectively acts as a learned implicit proxy for the theoretical $\alpha_l$ gating parameters. For analytical clarity to represent this ideal explicit scaling, we retain the following gating formulation: 
\begin{equation} \label{eq:meta_regulator} 
    [\alpha_1, \dots, \alpha_L] \;=\; \mathcal{M}_\psi(Z),
    \qquad \alpha_l \in [0, 1],
\end{equation}
where $\mathcal{M}_\psi$ denotes the full transformer encoder. In this conceptual formulation, $\alpha_l$ represents the effective influence that the attended representation of token $z_l$ exerts on the final output of $\mathcal{M}_\psi$, as governed by the self-attention operation in Equation~(\ref{eq:attention}). This influence shapes the magnitude of the gradient signal propagated back to stage $l$. The effective gradient update for layer $l$ is then expressed as
\begin{equation} \label{eq:gradient_modulation}
    \tilde{g}_l \;=\; \alpha_l \cdot \frac{\partial \mathcal{L}}{\partial
    \theta_l},
    \qquad
    \theta_l^{(t+1)} \;=\; \theta_l^{(t)} \,-\, \eta\, \tilde{g}_l.
\end{equation}
While Equation~(\ref{eq:gradient_modulation}) shows an explicit scalar multiplication, in practice, this scaling is naturally executed by the chain rule passing through the attention mechanism. The key assumptions underlying this formulation are as follows: (i) a global view across $\{z_l\}$ contains sufficient signals for estimating the relative learning need at each depth; (ii) $\mathcal{M}_\psi$ is end-to-end differentiable so that $\psi$ is learned jointly with $\{\theta_l\}$ from the task loss;  (iii) the modulation is adaptive per input batch, so $\alpha_l$ varies with both the sample and the training stage. Under this view, residual connections correspond to the degenerate case $\alpha_l = 1$ for all $l$ and $t$, and GradAttn generalizes this to a learned, input-conditioned gradient routing policy.

\subsection{Architecture Design}
\label{subsec:arch design}

\begin{figure}
\centering
    \includegraphics[width=\textwidth]{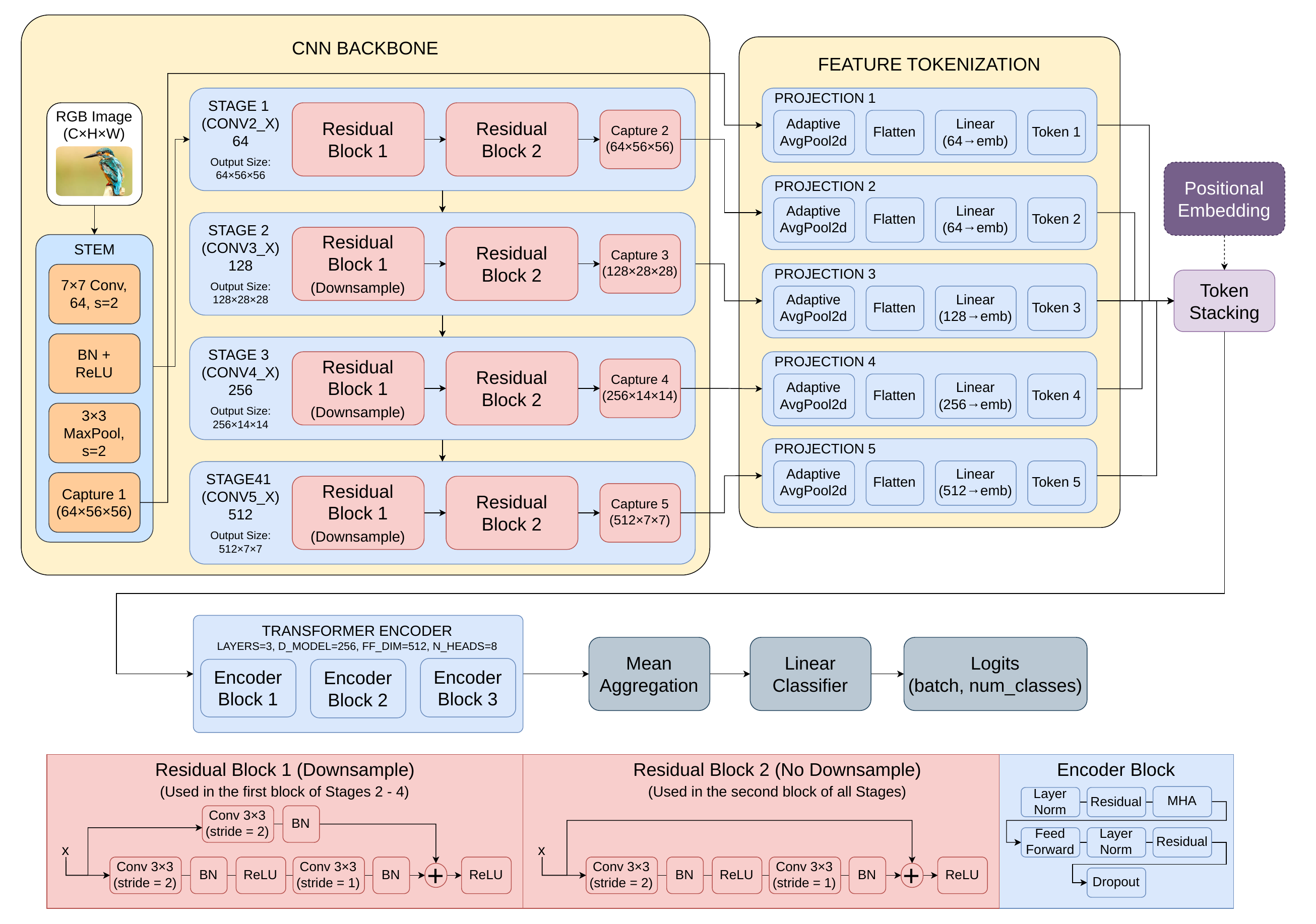}
    \caption{Proposed attention-controlled gradient flow architecture [GradAttn].}
    \label{fig:gradattn_arch}
\end{figure}

Our GradAttn framework (Figure~\ref{fig:gradattn_arch}) replaces ResNet's fixed residual connections with attention-controlled gradient pathways. The backbone follows ResNet-18's convolutional structure but removes all skip connections. Instead, we extract features at five strategic depths: after initial max pooling and after each of the five convolutional stages, yielding feature maps $\{f_1, f_2, f_3, f_4, f_5\}$.

Each feature map $f_i \in \mathbb{R}^{c_i \times h_i \times w_i}$, where $c_i$, $h_i$, and $w_i$ denote the number of channels, height, and width at extraction point $i$, respectively, undergoes global average pooling and linear projection into a common embedding dimension $d$:
\begin{equation} \label{eq:projection}
z_i = W_p^i \cdot \text{Pool}(f_i)
\end{equation}
where $\text{Pool}(\cdot): \mathbb{R}^{c_i \times h_i \times w_i} \rightarrow \mathbb{R}^{c_i}$ denotes global average pooling that spatially aggregates each channel into a scalar, and $W_p^i \in \mathbb{R}^{d \times c_i}$ projects the pooled features into the shared embedding space, yielding $z_i \in \mathbb{R}^{d}$.

The five extraction points correspond to ResNet-18's architectural structure: after initial max pooling and after each of the four convolutional stages (every two basic blocks). This configuration balances computational efficiency with comprehensive gradient flow coverage. Increasing to per-block extraction (10 points) would significantly increase memory overhead without substantial benefit, as our Gradient Health Score analysis (Section~\ref{sec:ghs}) shows minimal vanishing gradients within two consecutive blocks. Reducing below five points would skip entire stages, essentially creating a vanilla CNN. Thus, five points represent the optimal balance for multi-scale feature integration.

The resulting token sequence $Z = [z_1, z_2, z_3, z_4, z_5]$ enters a transformer encoder with $L$ attention layers (we use $L=3$ attention layers with $8$ heads, an embedding dimension of $d=256$, and standard feedforward networks of $f\_dim=512$ with layer normalization).

Multi-head self-attention computes queries, keys, and values as follows:
\begin{equation}
Q = ZW_Q, \quad K = ZW_K, \quad V = ZW_V
\end{equation}
The attention operation dynamically weights feature contributions:
\begin{equation} \label{eq:attention}
\text{Attention}(Q,K,V) = \text{softmax}\left(\frac{QK^T}{\sqrt{d}}\right)V
\end{equation}
This mechanism learns adaptive gradient pathways that dynamically shape learned representations; unlike ResNet's uniform short circuits, attention can emphasize deep semantic features for complex inputs while prioritizing shallow texture representations for simpler patterns.

\subsection{Positional Encoding Variants}

We evaluate three variants to understand spatial encoding requirements:

\textbf{No PE:} Relies solely on CNN-derived spatial structure.

\textbf{Learnable PE:} Adds trainable position embeddings:
\begin{equation}
z_i^{PE} = z_i + p_i, \quad p_i \in {R}^d
\end{equation}
where $p_i$ is a trainable parameter adapted during training. This allows domain-specific positional cues while maintaining the same transformer pipeline as the No PE variant.

\textbf{RoPE:} Relative positional information is incorporated directly into the attention mechanism by rotating queries and keys:
\begin{equation}
Q_i^{\text{RoPE}} = R_{\theta(i)}(Q_i), \quad K_i^{\text{RoPE}} = R_{\theta(i)}(K_i)
\end{equation}
where $R_{\theta(i)}$ is a rotation operator parameterized by position $i$. This design integrates spatial relationships natively into the dot-product attention computation.

\section{Experiments and Analysis} \label{section:results}

We present a comprehensive evaluation of GradAttn across eight diverse datasets, examining classification performance, gradient flow characteristics, model calibration, and computational efficiency. Our analysis reveals the dataset-specific advantages of attention-controlled gradient pathways and provides insights into when and why learnable gradient routing outperforms fixed residual connections.

\subsection{Training Protocol}

All models are trained using the Adam optimizer with a fixed random seed for reproducibility. Full training and transformer hyperparameters are summarized in Table~\ref{tab:training_hyperparams}. Learning rate scheduling follows ReduceLROnPlateau monitoring validation accuracy, with training capped at 100 epochs and early stopping restoring the best observed weights. Dataset-specific preprocessing and augmentation strategies are detailed in Table~\ref{tab:data_processing}; augmentation is applied exclusively to training splits, with validation and test splits receiving only resizing and normalization. Both GradAttn and ResNet-18 use identical preprocessing pipelines throughout all experiments, ensuring that performance differences are attributable solely to architectural design rather than data handling.

\begin{table}
\caption{Training hyperparameters}
\label{tab:training_hyperparams}
\centering
\begin{tabular}{llll}
\toprule
\textbf{Hyperparameter} & \textbf{Value} & \textbf{Hyperparameter} & \textbf{Value} \\
\midrule
Optimizer & Adam & Early Stopping Monitor & Val Accuracy \\
Learning Rate & 1 $\times$ 10$^{-3}$ & Early Stopping Patience & 7 epochs \\
Weight Decay & 1 $\times$ 10$^{-4}$ & Best Weight Restoration & Enabled \\
Batch Size & 128 & LR Scheduler & ReduceLROnPlateau \\
Maximum Epochs & 100 & LR Patience & 3 epochs \\
Random Seed & 42 & LR Reduction Factor & 0.2 \\
& & Minimum LR & 1 $\times$ 10$^{-7}$ \\
\midrule
\multicolumn{4}{l}{Transformer Encoder Settings} \\
\midrule
Embedding Dim & 256 & Feedforward Dim & 512 \\
Attention Heads & 8 & Dropout & 0.1 \\
Encoder Layers & 3 & & \\
\bottomrule
\end{tabular}
\end{table}

\subsection{Evaluation Datasets}

\begin{figure}
\centering
\includegraphics[width=\textwidth]{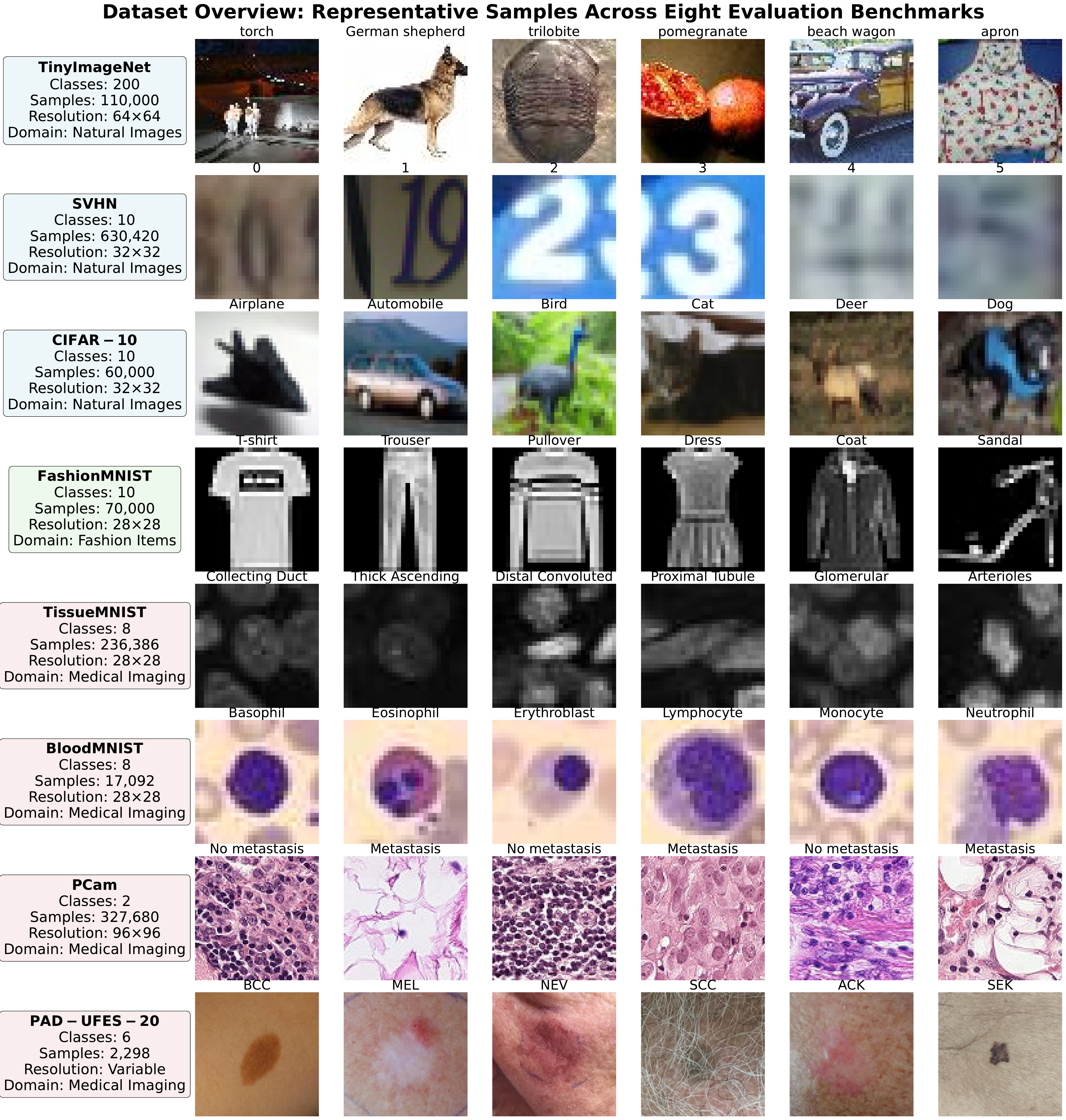}
\caption{Dataset overview and representative samples.}
\label{fig:datasets_fig}
\end{figure}

We carry out testing on eight diverse benchmarks: Tiny ImageNet (primary)~\citep{le2015tiny}, CIFAR-10~\citep{krizhevsky2009learning}, SVHN~\citep{netzer2011reading}, FashionMNIST~\citep{xiao2017fashion}, and four medical datasets (TissueMNIST~\citep{yang2023medmnist}, BloodMNIST~\citep{yang2023medmnist}, PCam~\citep{veeling2018rotation}, and PAD-UFES-20~\citep{pacheco2020pad}). This spans natural images, structured recognition, and specialized medical imaging to validate generalizability across domains (Figure~\ref{fig:datasets_fig}).
Tiny ImageNet, CIFAR-10, SVHN, FashionMNIST, and PAD-UFES-20 use a 70/15/15~train/val/test split. TissueMNIST and BloodMNIST use a 70/10/20 split to provide a larger test set given their medical imaging context. PCam uses an 80/10/10 split to maximize training data for the large-scale binary classification task.

\subsection{Classification Performance}

Table~\ref{tab:accuracy} summarizes the Top-1 accuracy results across all datasets. GradAttn variants outperform ResNet-18 on five of eight datasets, with performance gains strongly correlated with task complexity and the need for multi-scale feature integration. The RoPE variant achieves the highest accuracy on Tiny ImageNet (37.67\%, +4.65\% over ResNet-18) and SVHN (98.15\%, +0.17\%), while Learnable PE excels on FashionMNIST (75.18\%, +11.07\%) and matches ResNet-18 on TissueMNIST (69.88\%).

\begin{table}
\caption{Data preprocessing and augmentation settings per dataset (training split).}
\label{tab:data_processing}
\centering
\begin{tabular}{lll}
\toprule
\textbf{Dataset} & \textbf{Preprocessing Applied (Training)} & \textbf{Normalization (Mean / std)} \\

\midrule

Tiny ImageNet & None & \makecell[l]{[0.485, 0.456, 0.406] / \\{}[0.229, 0.224, 0.225]} \\

\midrule

CIFAR-10 & \makecell[l]{RandomHorizontalFlip (\emph{p} = 0.5),\\RandomCrop (32, \emph{pad} = 4)} & \makecell[l]{[0.4914, 0.4822, 0.4465] / \\{}[0.2023, 0.1994, 0.2010]} \\

\midrule

SVHN & None & \makecell[l]{[0.4377, 0.4438, 0.4728] / \\{}[0.1980, 0.2010, 0.1970]} \\

\midrule

FashionMNIST & \makecell[l]{RandomHorizontalFlip (\emph{p} = 0.5),\\RandomRotation (10\textdegree), \\RandomAffine (\emph{translate} = 0.1)} & [0.2860] / \\{}[0.3530] \\

\midrule

TissueMNIST & \makecell[l]{Resize (64 $\times$ 64), RandomRotation (10\textdegree), \\RandomHorizontalFlip (\emph{p} = 0.5), \\ColorJitter (\emph{b} = 0.1, \emph{c} = 0.1)} & \makecell[l]{[0.485, 0.456, 0.406] / \\{}[0.229, 0.224, 0.225]} \\

\midrule

BloodMNIST & \makecell[l]{Resize (64 $\times$ 64), RandomRotation (10\textdegree), \\RandomHorizontalFlip (\emph{p} = 0.5), \\ColorJitter (\emph{b} = 0.1, \emph{c} = 0.1)} & \makecell[l]{[0.485, 0.456, 0.406] / \\{}[0.229, 0.224, 0.225]} \\

\midrule

PCam & None & \makecell[l]{[0.485, 0.456, 0.406] / \\{}[0.229, 0.224, 0.225]} \\

\midrule

PAD-UFES-20 & \makecell[l]{Resize (224 $\times$ 224), \\RandomHorizontalFlip (\emph{p} = 0.5), \\RandomRotation (10\textdegree), \\ColorJitter (\emph{b} = 0.2, \emph{c} = 0.2, \emph{s} = 0.2, \emph{h} = 0.1), \\RandomResizedCrop (224, \emph{scale} = 0.8 - 1.0)} & \makecell[l]{[0.485, 0.456, 0.406] / \\{}[0.229, 0.224, 0.225]} \\

\bottomrule
\end{tabular}
\footnotesize{\\Note: No augmentation is applied to validation or test splits for any dataset.}
\end{table}

GradAttn underperforms ResNet-18 on three datasets: CIFAR-10 (by 2.94 - 5.61\%), PCam (by 0.89\%), and PAD-UFES-20 (by 4.91 - 7.80\%). On CIFAR-10, ResNet-18's strong convolutional inductive biases for low-resolution 32 $\times$ 32 images provide an advantage that multi-scale attention routing cannot overcome. On PCam, the binary metastasis detection task has simple decision boundaries where local convolutional features suffice. On PAD-UFES-20, the limited dataset size of only 2298 samples causes the transformer components to overfit.

\begin{table}
\caption{Top-1 accuracy comparison across datasets.}
\label{tab:accuracy}
\centering
\begin{tabular}{lcccc}
\toprule
\textbf{Dataset} & \textbf{ResNet-18} & \textbf{No PE} & \textbf{Learnable PE} & \textbf{RoPE}  \\
\midrule
Tiny ImageNet & 33.02\% & 35.89\% & 34.72\% & 37.67\% \\
SVHN         & 97.98\% & 98.09\% & 98.09\% & 98.15\%  \\
CIFAR-10     & 92.10\% & 88.68\% & 86.49\% & 89.16\%  \\
\midrule
FashionMNIST & 64.11\% & 66.70\% & 75.18\% & 62.40\%  \\
\midrule
TissueMNIST  & 69.88\% & 69.79\% & 69.88\% & 69.72\%  \\
BloodMNIST   & 95.97\% & 96.23\% & 94.48\% & 95.59\%  \\
PCam         & 80.71\% & 79.65\% & 79.74\% & 79.82\%  \\
PAD-UFES-20  & 55.49\% & 50.29\% & 50.58\% & 47.69\%  \\
\bottomrule
\end{tabular}
\end{table}

\subsubsection{Extended Evaluation}

Beyond Top-1 accuracy, we evaluate Top-3 and Top-5 accuracy, macro-averaged F1 scores, and the Expected Calibration Error (ECE) to assess ranking quality and prediction reliability. Table~\ref{tab:extended_metrics} presents these metrics for representative datasets where GradAttn demonstrates substantial improvements.

\begin{table}
\caption{Extended performance metrics.}
\centering
\label{tab:extended_metrics}
\begin{tabular}{lcccc}
\toprule
\textbf{Model} & \textbf{Top-3 Acc} & \textbf{Top-5 Acc} & \textbf{F1 Score} & \textbf{ECE} \\
\midrule
\multicolumn{5}{c}{Tiny ImageNet} \\
\midrule
ResNet-18 & 50.52\% & 58.62\% & 0.327 & 0.344 \\
RoPE & 56.45\% & 64.81\% & 0.373 & 0.197 \\
\midrule
\multicolumn{5}{c}{FashionMNIST} \\
\midrule
ResNet-18   & 92.21\% & 98.06\% & 0.589 & 0.193 \\
Learnable PE & 96.93\% & 99.30\% & 0.723 & 0.121 \\
\midrule
\multicolumn{5}{c}{TissueMNIST} \\
\midrule
ResNet-18 & 93.83\% & 98.55\% & 0.695 & 0.067 \\
Learnable PE & 93.92\% & 98.64\% & 0.696 & 0.039 \\
\bottomrule
\end{tabular}
\end{table}

The ECE reductions of 35 - 45\% across successful domains indicate that attention-controlled gradient flow improves not only accuracy but also calibration. The dynamic weighting mechanism produces more reliable confidence estimates by selectively emphasizing features relevant to each input, whereas uniform residual connections propagate all features equally regardless of their contribution to the prediction. 

The ECE is defined as follows: 
\begin{equation}
ECE = \sum_{m=1}^{M} \frac{|B_m|}{n} \Big| \text{acc}(B_m) - \text{conf}(B_m) \Big|
\end{equation}
where $M$ is the number of bins, $B_m$ represents the set of predictions in the $m$-th bin, $n$ is the total number of samples, $\text{acc}(B_m)$ is the average accuracy of bin $B_m$, and $\text{conf}(B_m)$ is the average predicted confidence of bin $B_m$.

\subsubsection{Comprehensive Performance Analysis}

Table~\ref{tab:comp_metrics} presents precision, recall, and F1 scores (both macro- and weighted averages) across all datasets and variants. These metrics reveal nuanced performance characteristics beyond simple accuracy measurements.

\begin{table}
\caption{Precision, recall, and F1 score across datasets.}
\label{tab:comp_metrics}
\centering
\begin{tabular}{lcccccc}
\toprule
\textbf{Model} & \makecell{\textbf{Precision}\\\textbf{(M)}} & \makecell{\textbf{Precision}\\\textbf{(W)}} & \makecell{\textbf{Recall}\\\textbf{(M)}} & \makecell{\textbf{Recall}\\\textbf{(W)}} & \makecell{\textbf{F1}\\\textbf{(M)}} & \makecell{\textbf{F1}\\\textbf{(W)}} \\
\midrule
\multicolumn{7}{c}{Tiny ImageNet} \\
\midrule
ResNet-18 & 0.328 & 0.331 & 0.330 & 0.330 & 0.327 & 0.329 \\
RoPE      & 0.371 & 0.374 & 0.377 & 0.377 & 0.371 & 0.373 \\
\midrule
\multicolumn{7}{c}{FashionMNIST} \\
\midrule
ResNet-18   & 0.662 & 0.662 & 0.636 & 0.641 & 0.585 & 0.589 \\
Learnable PE & 0.760 & 0.760 & 0.748 & 0.752 & 0.721 & 0.723 \\
\midrule
\multicolumn{7}{c}{TissueMNIST} \\
\midrule
ResNet-18 & 0.631 & 0.695 & 0.609 & 0.699 & 0.617 & 0.695 \\
Learnable PE & 0.632 & 0.696 & 0.613 & 0.699 & 0.618 & 0.695 \\
\bottomrule
\end{tabular}

\footnotesize{M: Macro-averaged; W: weighted-averaged across classes.}
\end{table}

For highly imbalanced datasets like Tiny ImageNet (200 classes with varying difficulty), macro-averaged metrics reveal that GradAttn (RoPE) achieves more balanced performance across classes, with macro-F1 score improvements of +0.044 over ResNet-18. The weighted metrics confirm that improvements are not merely driven by better performance on dominant classes but reflect genuine enhancement in handling diverse visual categories.

On FashionMNIST, Learnable PE achieves remarkable macro-F1 score gains (+0.136), indicating substantially improved recognition of challenging classes like pullovers and shirts that ResNet-18 frequently confuses. This suggests that dataset-specific positional adaptations enable the model to capture subtle texture and shape variations critical for fine-grained fashion recognition.

\subsection{Domain-Specific Patterns}

\subsubsection{Complex Natural Images}

On Tiny ImageNet and SVHN, RoPE consistently outperforms other variants. Tiny ImageNet contains highly diverse visual categories (200 classes spanning objects, animals, and scenes) with significant intra-class variations and cluttered backgrounds. RoPE's relative positional encoding effectively captures spatial relationships in these complex scenes, enabling the attention mechanism to weight features based on their geometric configuration. The +4.65\% accuracy gain on Tiny ImageNet demonstrates that explicitly modeling relative spatial structure benefits tasks requiring global context understanding.

SVHN (street view house numbers) presents a different challenge: Digits appear at varying scales and orientations with complex backgrounds. Despite the simpler task structure (10 classes), RoPE achieves +0.17\% improvement by better handling spatial transformations through rotation-equivariant positional encoding. The near-ceiling performance (98.15\%) indicates that attention-controlled gradients can approach optimal performance on well-structured recognition tasks.

CIFAR-10 represents a case where GradAttn variants consistently underperform ResNet-18 (92.10\%), with No PE achieving 88.68\%, Learnable PE achieving 86.49\%, and RoPE achieving 89.16\%. Unlike Tiny ImageNet and SVHN where multi-scale feature integration provides measurable benefits, CIFAR-10's low-resolution 32 $\times$ 32 images contain limited spatial complexity that does not warrant the adaptive gradient routing GradAttn provides. At this resolution, discriminative information is largely captured by local convolutional features within the early layers, making ResNet-18's fixed residual connections sufficient and the transformer's cross-scale attention redundant. Furthermore, the relatively small intra-class variation and well-separated decision boundaries in CIFAR-10 mean that shallow texture features alone adequately represent the ten object categories, leaving little room for the deep semantic weighting that attention-controlled gradients excel at. This suggests that GradAttn's benefits are contingent on sufficient input resolution and feature hierarchy complexity to justify the learnable gradient routing mechanism.

\subsubsection{Fashion and Texture Recognition}

FashionMNIST represents structured object recognition where shape and texture jointly determine categories. Learnable PE achieves the most substantial improvement (+11.07\%) by adapting positional encodings to clothing-specific patterns. Unlike natural images where spatial relationships follow universal geometric principles, fashion items exhibit dataset-specific structural regularities (e.g., shirts always have sleeves in particular positions, trousers have consistent leg structures). Learnable positional encodings capture these domain-specific spatial priors more effectively than either fixed CNN hierarchies (No PE) or universal relative encodings (RoPE).

\subsubsection{Medical Imaging}

Medical imaging datasets exhibit divergent patterns that illuminate when attention-controlled gradients provide advantages:

BloodMNIST (Blood Cell Classification): The No PE variant achieves the highest accuracy (96.23\%, +0.26\% over ResNet-18), suggesting that CNN-derived spatial hierarchies sufficiently encode the structure of microscopy images. Blood cells have consistent internal structure (nucleus and cytoplasm) with discrimination primarily based on morphological features at fixed scales. The strong performance without explicit positional encoding indicates that convolutional inductive biases already capture medically relevant spatial~patterns.

TissueMNIST (Kidney Tissue Classification): Learnable PE matches ResNet-18 in Top-1 accuracy but achieves superior Top-3 (+0.09\%) and Top-5 (+0.09\%) performance alongside substantially reduced ECE (0.039 vs. 0.067, $-$41.8\%). This pattern suggests that attention-controlled gradients improve confidence calibration and ranking quality even when final classification accuracy remains comparable. For medical applications where physicians review top-k predictions, improved ranking reliability provides clinical value beyond raw~accuracy.

PCam and PAD-UFES-20: ResNet-18 outperforms GradAttn variants on both datasets (+0.89\% on PCam, +4.91\% on PAD-UFES-20). PCam involves binary metastasis detection in histopathology patches with simple object boundaries; PAD-UFES-20 contains only 2298~samples across six skin lesion types. These results indicate two scenarios where attention-controlled gradients provide limited benefit: (1) tasks with simple decision boundaries where local convolutional features suffice and (2) small datasets where transformer components overfit due to insufficient training samples. The negative results validate that GradAttn's benefits depend on task complexity and dataset scale rather than universally improving upon residual connections.

\subsection{Training Dynamics and Convergences} \label{sec:train_dyn}

\begin{figure}
    \centering
    \begin{subfigure}{0.49\textwidth}
        \centering
        \includegraphics[width=\textwidth]{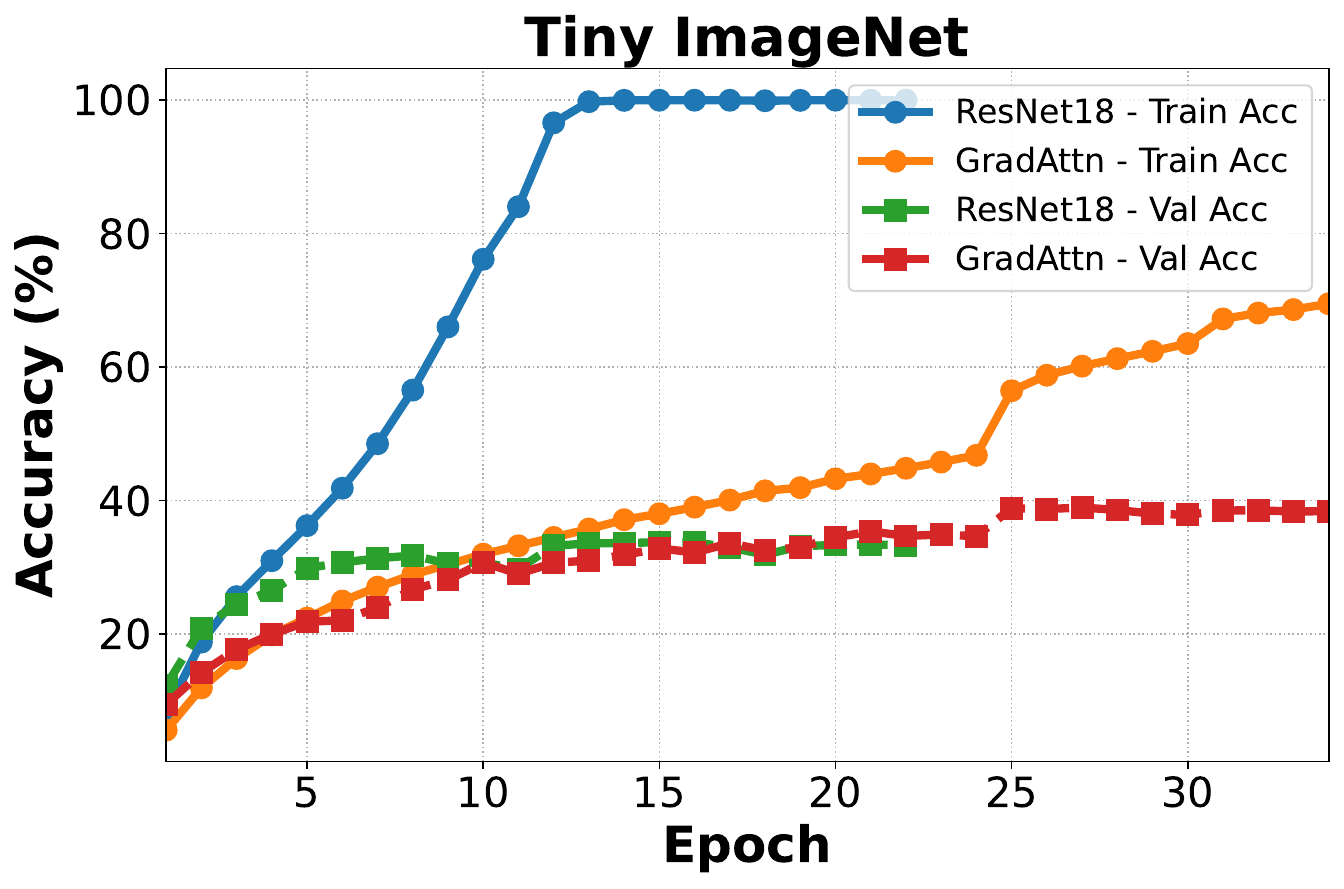}
        \caption{Resnet18 vs GradAttn (RoPE)}
        \label{fig:train_dynamics_tiny}
    \end{subfigure}
    \hfill
    \begin{subfigure}{0.49\textwidth}
        \centering
        \includegraphics[width=\textwidth]{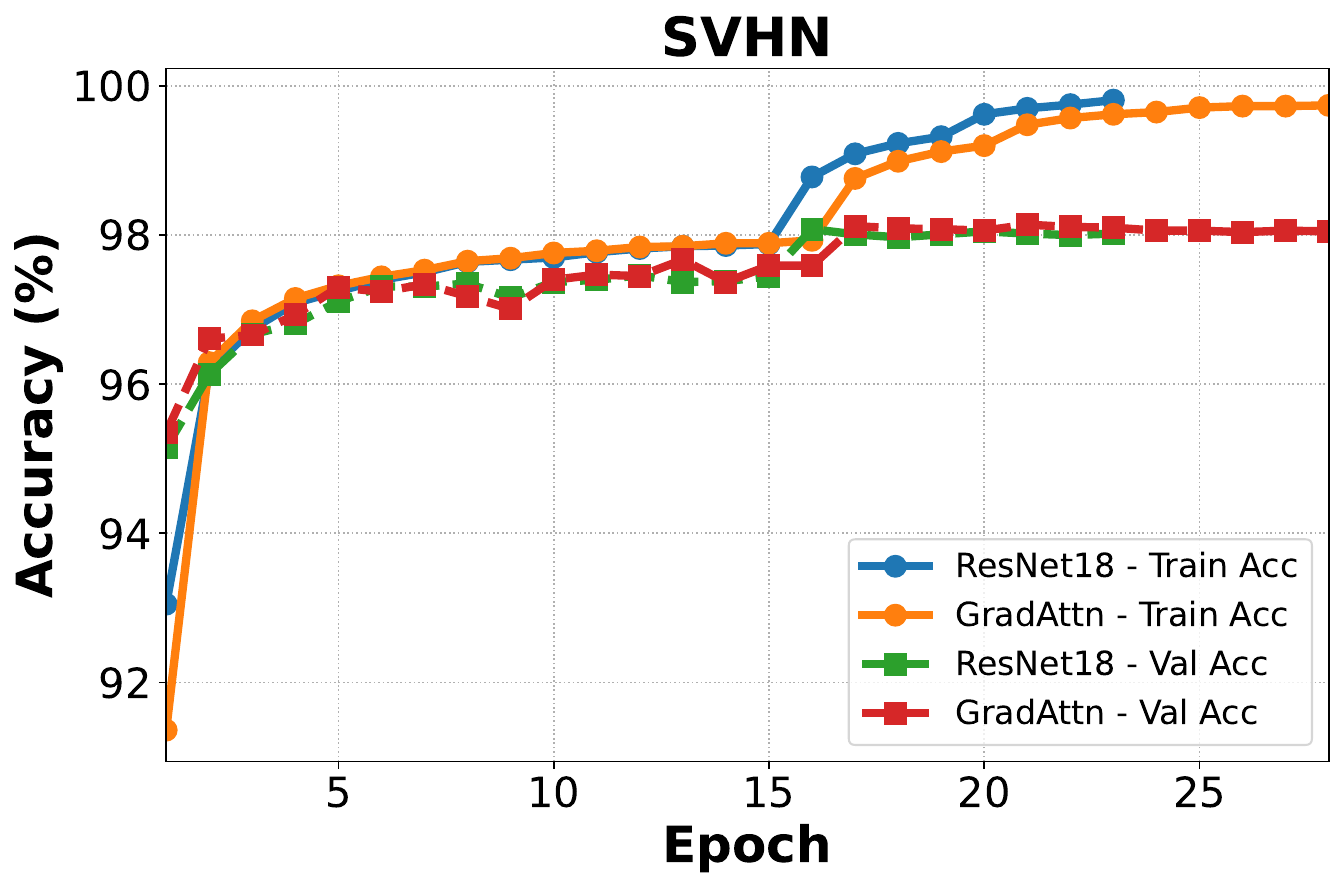}
        \caption{Resnet18 vs GradAttn (RoPE)}
        \label{fig:train_dynamics_svhn}
    \end{subfigure}
    
    \vspace{0.5cm}
    
    \begin{subfigure}{0.49\textwidth}
        \centering
        \includegraphics[width=\textwidth]{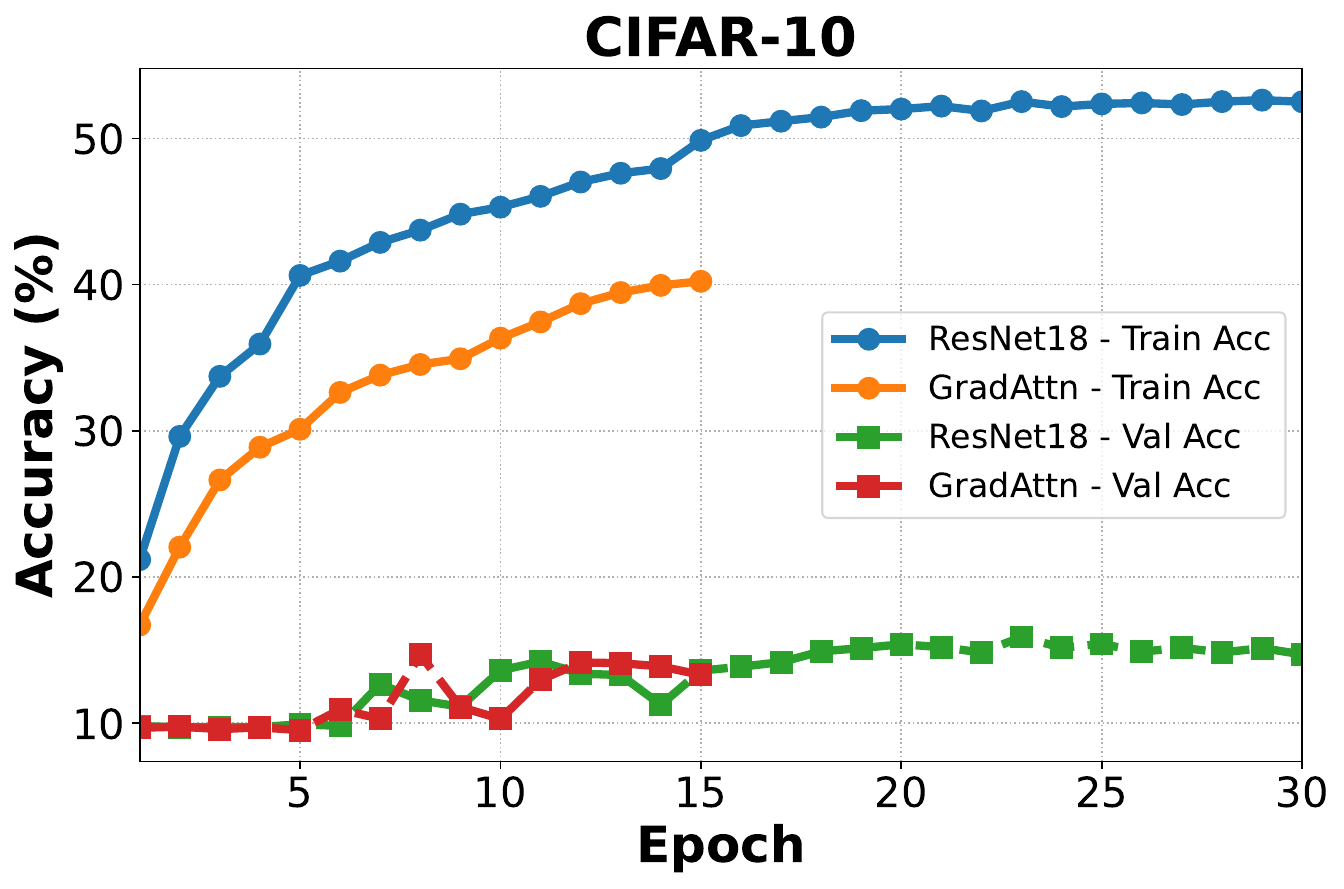}
        \caption{Resnet18 vs GradAttn (Learnable PE)}
        \label{fig:train_dynamics_cifar10}
    \end{subfigure}
    \hfill
    \begin{subfigure}{0.49\textwidth}
        \centering
        \includegraphics[width=\textwidth]{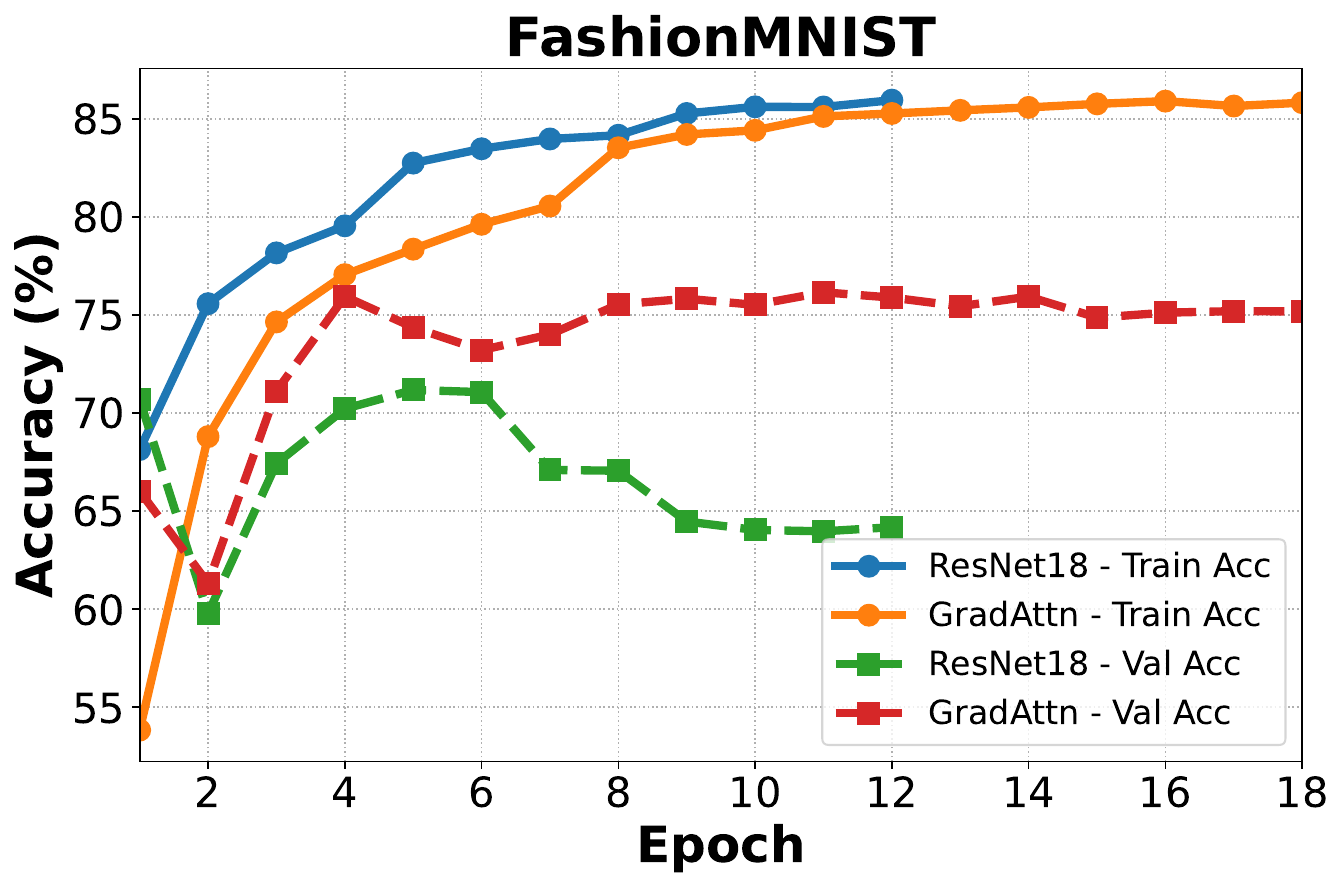}
        \caption{Resnet18 vs GradAttn (Learnable PE)}
        \label{fig:train_dynamics_fashion}
    \end{subfigure}
    
    \caption{Training dynamics comparison across Complex Natural Images and Texture Recognition datasets.}
    \label{fig:training_dynamics_complex}
\end{figure}

\begin{figure}
    \centering
    \begin{subfigure}{0.49\textwidth}
        \centering
        \includegraphics[width=\textwidth]{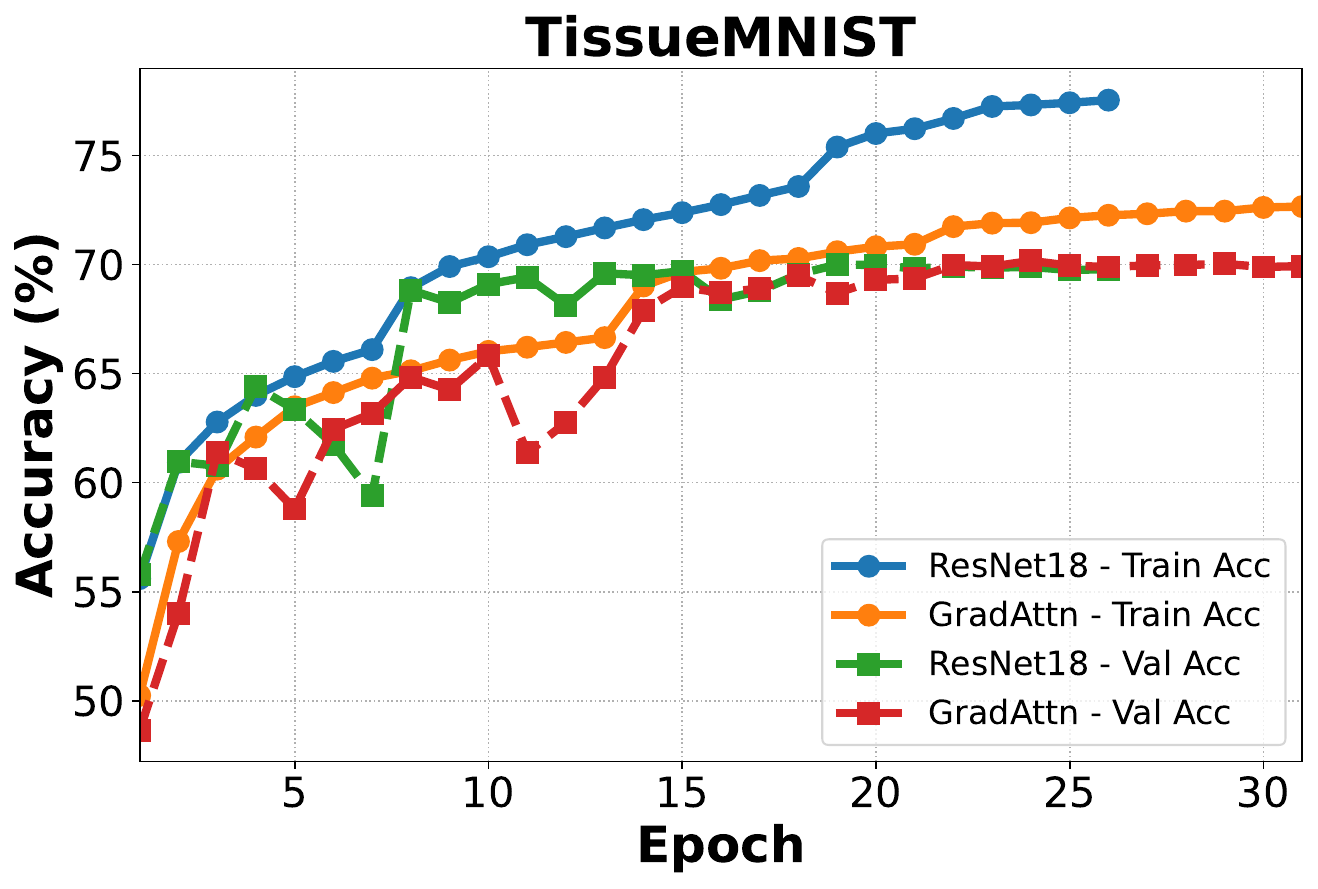}
        \caption{Resnet18 vs GradAttn (Learnable PE)}
        \label{fig:train_dynamics_tissue}
    \end{subfigure}
    \hfill
    \begin{subfigure}{0.49\textwidth}
        \centering
        \includegraphics[width=\textwidth]{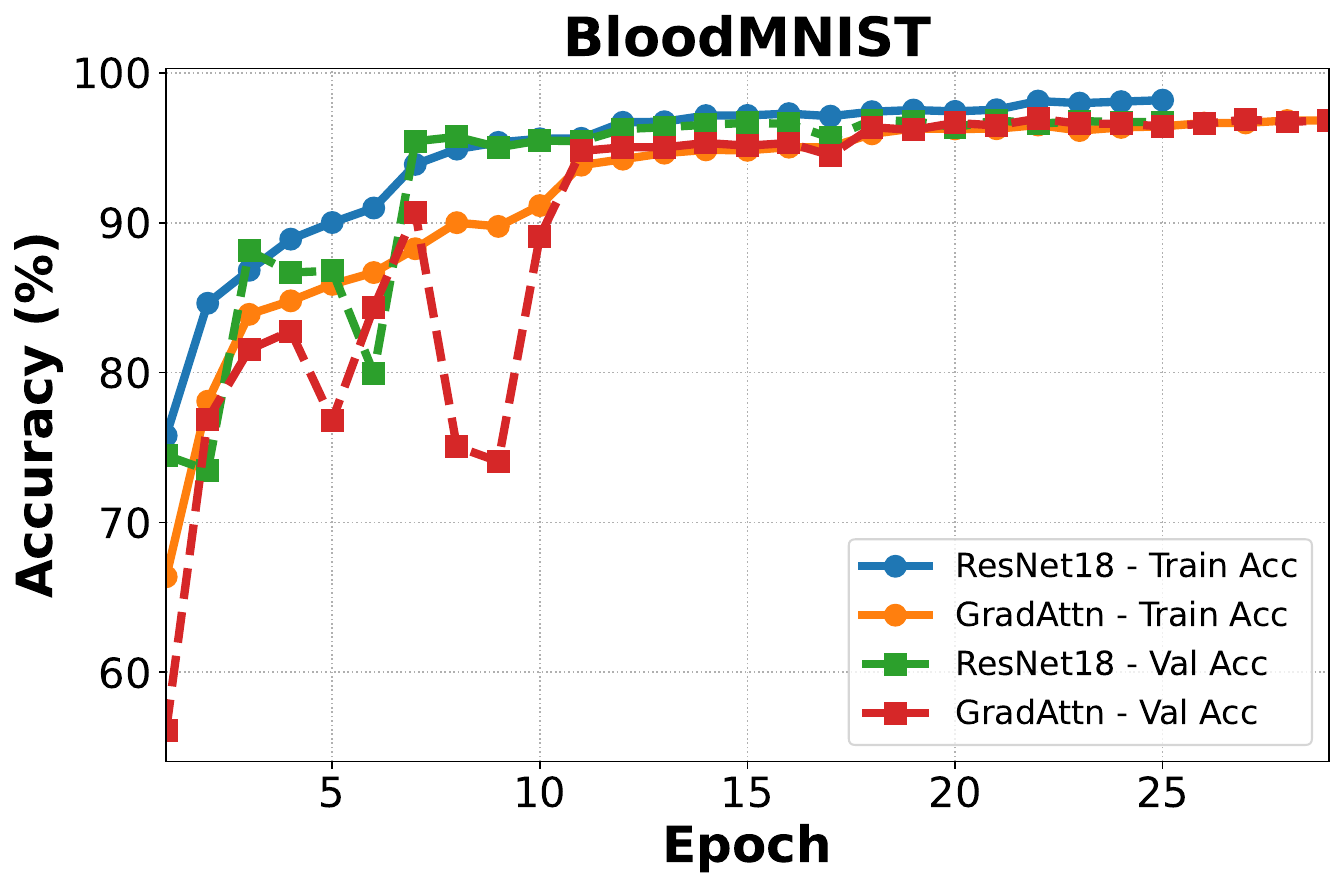}
        \caption{Resnet18 vs GradAttn (No PE)}
        \label{fig:train_dynamics_blood}
    \end{subfigure}
    
    \vspace{0.5cm}
    
    \begin{subfigure}{0.49\textwidth}
        \centering
        \includegraphics[width=\textwidth]{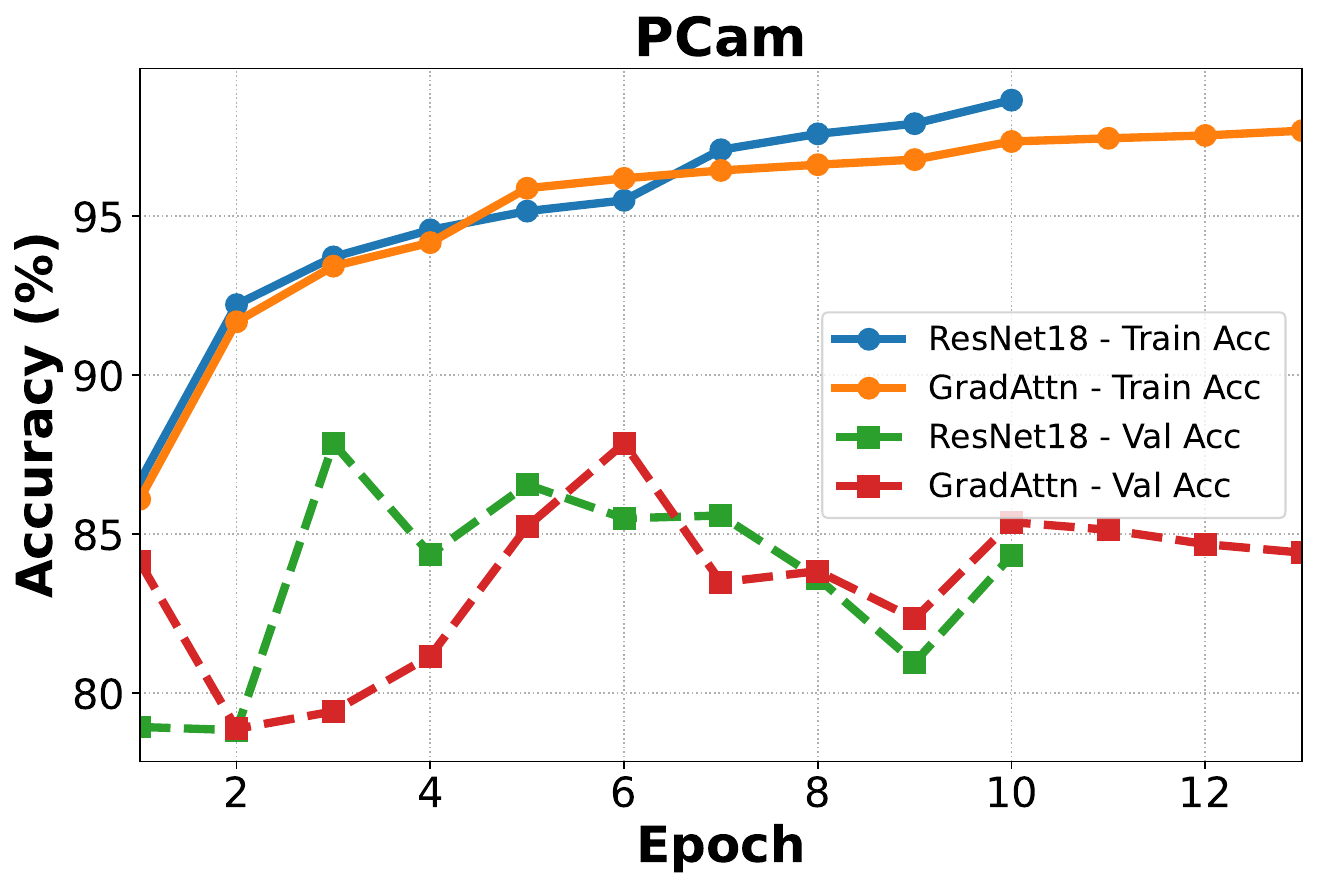}
        \caption{Resnet18 vs GradAttn (RoPE)}
        \label{fig:train_dynamics_pcam}
    \end{subfigure}
    \hfill
    \begin{subfigure}{0.49\textwidth}
        \centering
        \includegraphics[width=\textwidth]{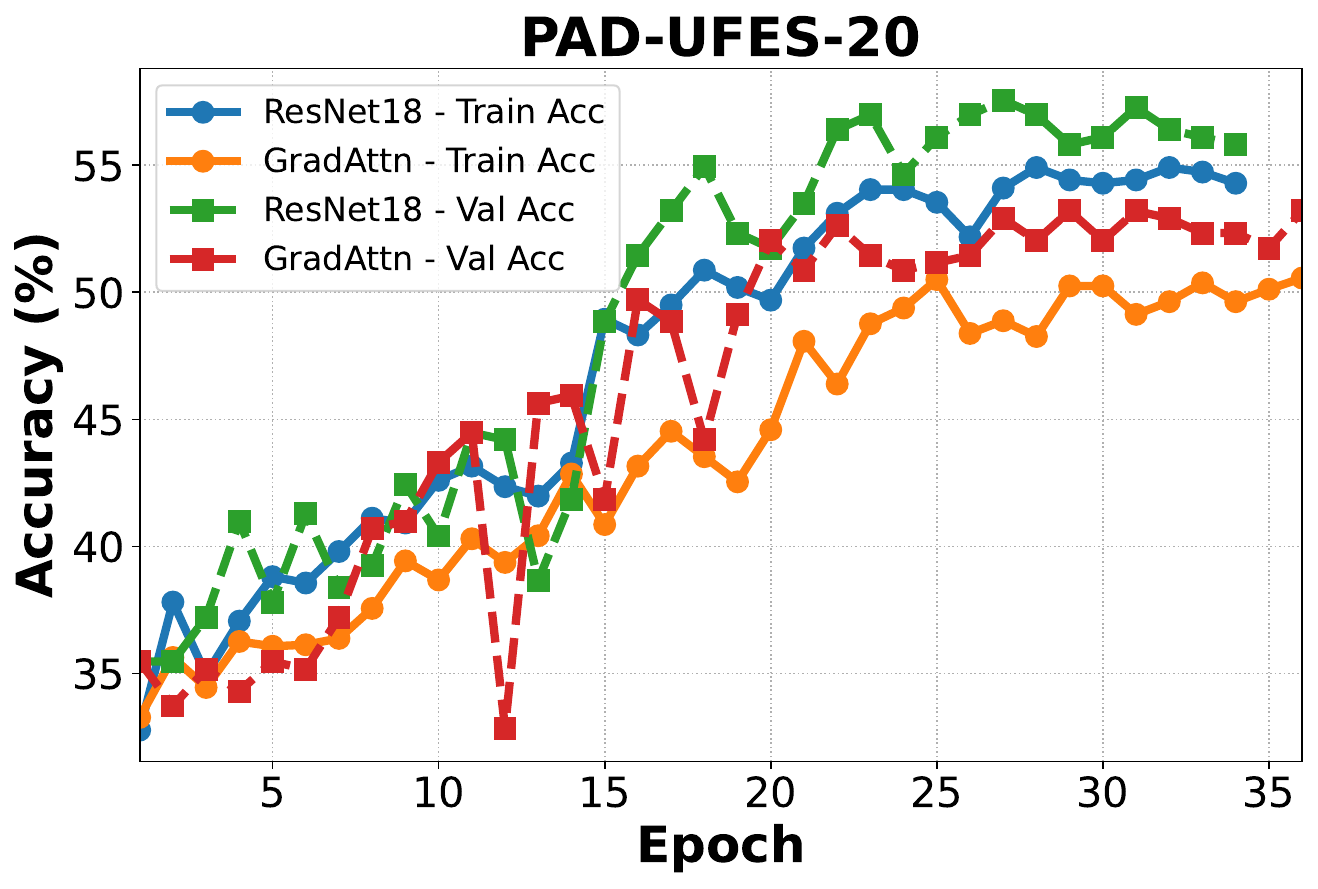}
        \caption{Resnet18 vs GradAttn (Learnable PE)}
        \label{fig:train_dynamics_pad}
    \end{subfigure}
    
    \caption{Training dynamics comparison across Medical datasets.}
    \label{fig:training_dynamics_medical}
\end{figure}

Figures~\ref{fig:training_dynamics_complex} and~\ref{fig:training_dynamics_medical} present training and validation accuracy curves comparing ResNet-18 with the best-performing GradAttn variant for each dataset. Contrary to expectations that attention mechanisms would accelerate convergence, we observe nuanced patterns where GradAttn variants often require more epochs to converge yet achieve superior final~performance.

\subsubsection{Convergence Speed and Training Requirements}

Across datasets, GradAttn models typically require more epochs to converge than ResNet-18. This does not reflect optimization inefficiency but rather the additional time needed for the attention mechanism to learn meaningful routing patterns. On FashionMNIST, for instance, the Learnable PE variant requires 11 epochs compared to ResNet-18’s five, which is a more than $2\times$ increase, yet this extended training yields a +11.07\% improvement in accuracy. This illustrates that the extra training time is compensated by substantially better representations that fixed skip connections cannot obtain. Overall, the modest increase in convergence time is justified by the consistent improvements in final performance.

\subsubsection{Optimization Landscape Complexity}

The extended training requirements for GradAttn reveal fundamental differences in optimization dynamics between fixed and learnable gradient pathways. We identify two~factors contributing to slower convergence:

Interdependent Component Optimization: GradAttn must simultaneously optimize CNN feature extraction, linear projections, and attention weights. Early in training, random attention patterns distribute gradients nearly uniformly across extraction points, providing minimal benefit over ResNet while adding optimization complexity. Only after the attention mechanism learns meaningful feature importance patterns (typically 15 - 25 epochs on Tiny ImageNet) does performance begin exceeding ResNet-18. This initial ``discovery phase'' explains the extended training time.

Non-Stationary Gradient Distributions: Unlike ResNet where gradient pathways remain constant, GradAttn's attention-controlled routing creates non-stationary optimization. As attention weights evolve, the effective learning rate for different layers changes dynamically, with layers receiving high attention experiencing larger gradient updates while low-attention layers adapt slowly. This adaptive behavior requires more iterations to reach equilibrium but produces better calibrated feature representations.

\subsubsection{Generalization Gap Analysis}

Table~\ref{tab:convergence_gap} quantifies the train-validation accuracy gap at convergence, revealing overfitting tendencies across architectures.

\begin{table}
\caption{Train-validation accuracy gap at convergence.}
\label{tab:convergence_gap}
\centering
\begin{tabular}{lcccc}
\toprule
\textbf{Dataset} & \textbf{ResNet-18} & \textbf{No PE} & \textbf{Learnable PE} & \textbf{RoPE}  \\
\midrule
Tiny ImageNet & 65.77\% & 16.24\% & 14.30\% & 25.38\% \\
SVHN          & 0.70\% & 0.94\% & 1.29\% & 1.34\% \\
CIFAR-10      & 6.52\% & 5.47\% & 3.16\% & 6.62\% \\
\midrule
FashionMNIST  & 11.57\% & 10.85\% & 8.97\% & 10.96\% \\
\midrule
TissueMNIST   & 5.39\% & 1.31\% & 1.75\% & 1.58\% \\
BloodMNIST    & 0.57\% & 0.43\% & 0.63\% & 0.15\% \\
PCam          & 5.86\% & 7.37\% & 7.31\% & 8.33\% \\
PAD-UFES-20   & 3.46\% & 3.27\% & 2.95\% & 4.21\% \\
\bottomrule
\end{tabular}

\footnotesize{\textit{Note.} $\text{Gap} = |\text{Train Accuracy}-\text{Validation Accuracy}|$}.
\end{table}

\vspace{-4pt}
GradAttn variants dramatically reduce overfitting on complex datasets where they achieve superior test accuracy. Most notably, on Tiny ImageNet, Learnable PE reduces the generalization gap from 65.77\% (ResNet-18) to 14.30\%, a 78.2\% reduction in overfitting. This massive improvement indicates that ResNet-18 severely overfits the training data, while attention-controlled gradients learn substantially more generalizable representations. The RoPE and No PE variants also achieve large reductions (61.4\% and 75.3\%, respectively), confirming that dynamic feature weighting inherently regularizes learning.

On FashionMNIST, Learnable PE reduces overfitting by 22.5\% (11.57\% to 8.97\%), while on TissueMNIST, No PE achieves a 75.7\% reduction (5.39\% to 1.31\%). These results suggest that attention-controlled gradients act as an adaptive regularizer, where less relevant features contribute minimally to gradients during backpropagation, preventing the model from memorizing spurious correlations in the training set.

CIFAR-10 shows a balanced generalization pattern, with ResNet-18 exhibiting a 6.52\%~gap and GradAttn variants showing comparable or smaller gaps (3.16 - 6.62\%). Despite Learnable PE achieving the tightest gap at 3.16\%, ResNet-18 maintains superior test accuracy at 92.10\%, indicating that GradAttn variants generalize more consistently but from a lower performance ceiling due to ResNet-18's stronger convolutional inductive biases for low-resolution inputs.

On medical imaging datasets, the pattern is nuanced. For TissueMNIST and BloodMNIST, GradAttn variants achieve minimal generalization gaps (0.15 - 1.75\%), substantially lower than ResNet-18 (5.39\% and 0.57\%, respectively), indicating excellent generalization. However, on PCam, ResNet-18 maintains a smaller gap (5.86\%) compared to GradAttn variants (7.31 - 8.33\%), aligning with our earlier finding that PCam's binary classification with simple decision boundaries favors ResNet's convolutional inductive biases. On PAD-UFES-20, all models exhibit relatively small gaps (2.95 - 4.21\%), with minimal differences between architectures, suggesting that the primary challenge is insufficient training data (only 2298 samples) rather than overfitting.

These results establish attention-controlled gradients as a powerful implicit regularizer for complex, large-scale datasets where multi-scale feature integration benefits from adaptive selection. The regularization effect emerges naturally from the attention mechanism's learned feature weighting rather than requiring explicit regularization techniques.

\subsection{Gradient Flow Analysis} \label{sec:ghs}
The Gradient Health Score is defined as follows:
\begin{equation}
GHS = \frac{N_{healthy}}{N_{total}}
\end{equation}
where $N_{healthy}$ represents layers with gradient norms in the range $[10^{-6}, 10]$, and $N_{total}$ is the total number of analyzed layers. $\text{GHS} = 1.0$ indicates perfect gradient stability, with all layers maintaining healthy gradient magnitudes, while $\text{GHS} < 1.0$ reveals the fraction of layers experiencing vanishing ($\text{gradient\_norm} < 10^{-6}$) or exploding ($\text{gradient\_norm} > 10$) gradients. This composite metric quantifies overall network gradient flow quality, where controlled instabilities ($0.8 < \text{GHS} < 1.0$) can coincide with improved generalization, as observed in 
our attention variants.

We monitored gradient health during testing using normalized stability metrics across all eight datasets. While ResNet-18 maintained perfect stability (GHS = 1.0) across experiments, attention variants consistently introduced controlled instabilities that often coincided with improved generalization. The No PE variant exhibited minimal gradient decay across most datasets, with generally stable training dynamics (e.g., GHS = 0.914 on FashionMNIST). The Learnable PE variant experienced localized vanishing gradients in several layers, yet it achieved the best performance on FashionMNIST (+11.07\%) and matched ResNet-18 on TissueMNIST while substantially improving calibration (ECE reduced by 41.8\%). Similarly, RoPE introduced controlled instability across a subset of layers while still achieving the highest accuracy on Tiny ImageNet and SVHN among all variants. Table~\ref{tab:gradient_flow_analysis} provides a concrete illustration of these gradient dynamics for the Tiny ImageNet dataset, where RoPE attains GHS = 0.732 with vanishing gradients in eight layers, yet it outperforms the perfectly stable ResNet-18 by +4.65\% in Top-1 accuracy.

This suggests that perfect gradient stability may not be optimal, i.e., controlled attention-induced redistribution can enhance feature learning despite minor gradient anomalies.

\begin{table}

\caption{Gradient flow analysis results on Tiny ImageNet.}
\label{tab:gradient_flow_analysis}

\centering
\begin{tabular}{lcccccccc}

\toprule

\multirow{3}{*}{\textbf{Model}} & \multicolumn{3}{c}{\textbf{GHS}} &
\multirow{3}{*}{\makecell{\textbf{V/E}\\\textbf{Gradients}}} &
\multirow{3}{*}{\makecell{\textbf{Avg}\\\textbf{Norm}}} &
\multirow{3}{*}{\textbf{Range}} &
\multirow{3}{*}{\textbf{Std}} \\
\cmidrule(lr){2-4}
& \footnotesize \textbf{\makecell{[1 $\times$ 10$^{-7}$,\\1 $\times$ 10$^{2}$]}} 
& \footnotesize \textbf{\makecell{[1 $\times$ 10$^{-6}$,\\1 $\times$ 10$^{1}$]}}
& \footnotesize \textbf{\makecell{[1 $\times$ 10$^{-5}$,\\1 $\times$ 10$^{0}$]}} & \\

\midrule

ResNet-18 
& 1.00 & 1.00 & 1.00 
& None 
& 0.153
& {0.022--0.370} 
& 0.106 \\

\midrule

\makecell[l]{GradAttn\\(No PE)} 
& 0.80 & 0.80 & 0.80
& \makecell{Vanishing\\(4 layers)} 
& 0.085
& {0.000--0.610} 
& 0.106 \\

\midrule

\makecell[l]{GradAttn\\(Learnable PE)} 
& 0.80 & 0.80 & 0.80 
& \makecell{Vanishing\\(4 layers)} 
& 0.078
& {0.000--0.336} 
& 0.070 \\

\midrule

\makecell[l]{GradAttn\\(RoPE)} 
& 0.73 & 0.73 & 0.73 
& \makecell{Vanishing\\(8 layers)} 
& 0.117 
& {0.000--0.957} 
& 0.199 \\

\bottomrule

\end{tabular}

\footnotesize{GHS = Gradient Health Score; V/E = vanishing/exploding. GHS values are consistent across all tested threshold\\ranges, confirming robustness of the metric.}
\end{table}

\subsubsection{Controlled Instability and Generalization}
The relationship between gradient stability and performance challenges conventional deep learning theory. On Tiny ImageNet, RoPE achieves the highest accuracy (37.67\%) despite exhibiting 0.732, with eight layers experiencing vanishing gradients. Learnable PE on FashionMNIST attains a +11.07\% improvement with GHS = 0.914. These results suggest that attention-induced gradient redistribution creates beneficial training dynamics where not all layers require uniform gradient flow.

We hypothesize that controlled instabilities enable the attention mechanism to effectively ``prune'' less relevant gradient pathways during training. Layers experiencing occasional vanishing gradients contribute minimally to feature learning, allowing the network to focus representational capacity on extraction points that are most critical for the task. This differs fundamentally from traditional vanishing gradient problems where poor initialization or activation functions cause universal gradient decay; attention-controlled vanishing is selective and input-dependent.

\begin{figure}
\includegraphics[width=\linewidth]{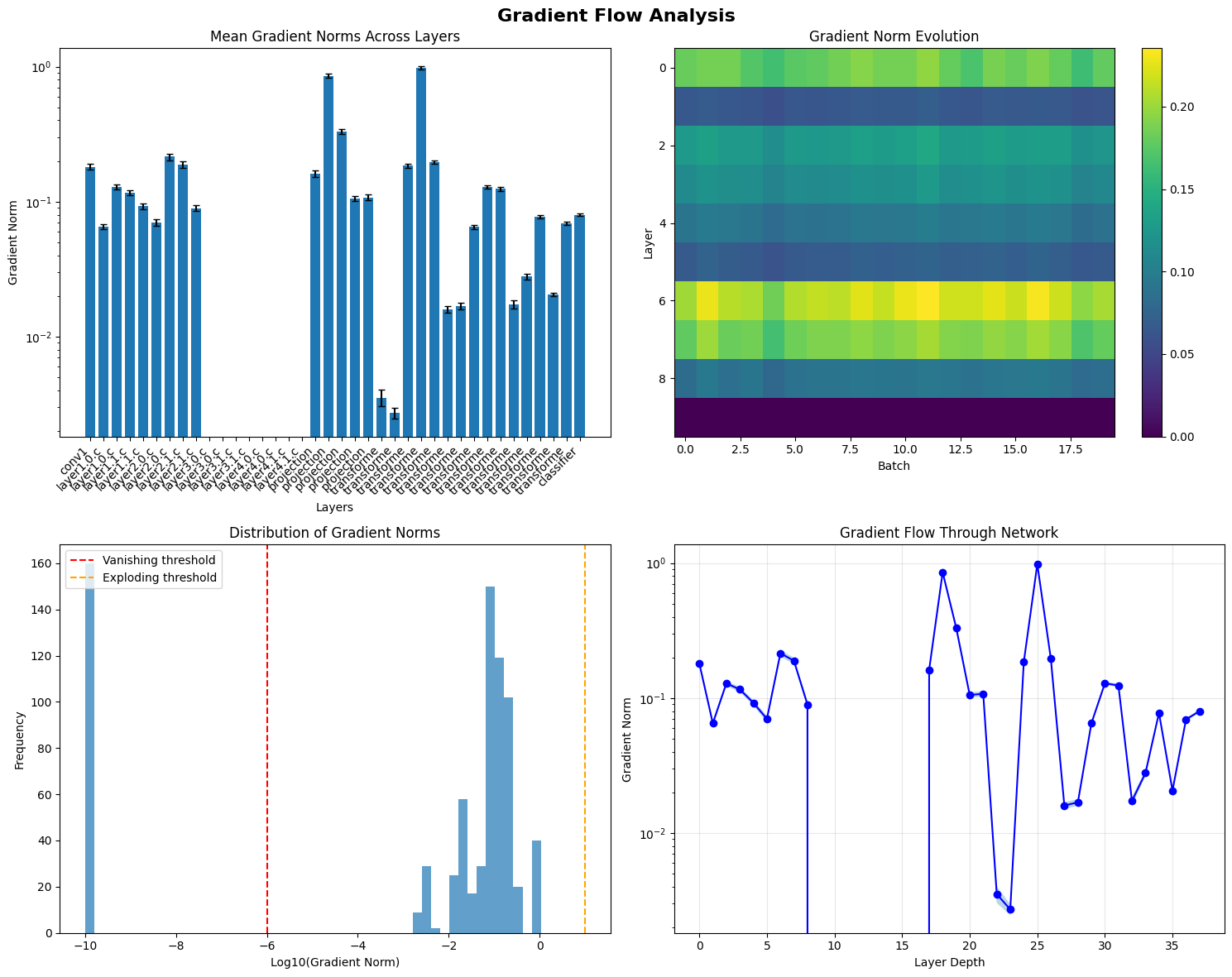}
\caption{Gradient dynamics of GradAttn (RoPE) on Tiny ImageNet dataset.}
\label{fig:gradients_tinyimagenet}
\end{figure}

Figure~\ref{fig:gradients_tinyimagenet} visualizes layer-wise gradient norm distributions, revealing that vanishing gradients in GradAttn models occur selectively in mid-depth layers, while shallow and deep layers maintain healthy gradients. This pattern indicates that attention learns to bypass intermediate feature hierarchies when they provide redundant information, creating direct pathways from shallow texture features and deep semantic representations to the loss~function.

\subsection{Parameter and Computational Efficiency}
Despite transformer layers, the models maintained competitive efficiency. While ResNet-18 has $\sim$11.2M parameters, attention variants added only $\sim$1.6M parameters ($\approx$14.3\% increase).
Table~\ref{tab:training_time} reports the average training time per epoch across all datasets and model variants. Contrary to expectations, GradAttn variants do not consistently incur higher per-epoch computational cost than ResNet-18, and in several cases, they achieve lower epoch times, notably on Tiny ImageNet, TissueMNIST, and PCam. This suggests that the attention-controlled gradient routing does not introduce significant computational overhead per epoch and that the extended total training time observed for GradAttn variants is attributable primarily to the greater number of epochs required for convergence rather than increased per-epoch cost.

\begin{table}
\caption{Average training time per epoch (seconds) across datasets.}
\label{tab:training_time}
\centering
\begin{tabular}{lcccc}
\toprule
\textbf{Dataset} & \textbf{ResNet-18} & \textbf{No PE} & \textbf{Learnable PE} & \textbf{RoPE} \\
\midrule
Tiny ImageNet  & 50.90  & 56.53  & 58.17  & 58.59  \\
SVHN           & 174.61 & 200.67 & 205.52 & 203.86 \\
CIFAR-10       & 42.03  & 40.45  & 40.50  & 40.84  \\
\midrule
FashionMNIST   & 36.33  & 37.42  & 37.50  & 38.94  \\
\midrule
TissueMNIST    & 142.00 & 119.43 & 152.71 & 158.90 \\
BloodMNIST     & 12.52  & 11.86  & 12.97  & 13.28  \\
PCam           & 209.20 & 198.92 & 197.31 & 198.15 \\
PAD-UFES-20    & 77.91  & 79.29  & 76.89  & 78.75 \\
\bottomrule
\end{tabular}

\footnotesize{Average time per epoch is computed over all epochs before early stopping.}
\end{table}

\subsection{Comparison with Attention-Augmented Baselines}

To further contextualise GradAttn's contributions beyond a plain ResNet-18 baseline, we compare against ResNet-18 augmented with established channel and spatial attention modules: Squeeze-and-Excitation (SE) Networks ~\cite{hu2018squeeze} and the Convolutional Block Attention Module (CBAM)~\cite{woo2018cbam}. These modules apply attention within individual residual blocks for feature recalibration, representing a natural intermediate between fixed residual connections and GradAttn's inter-layer gradient routing. Table~\ref{tab:se_cbam} presents Top-1 accuracy on Tiny ImageNet, SVHN, and FashionMNIST, chosen as representative datasets spanning complex natural images, structured digit recognition, and texture-based fashion recognition, respectively.

\begin{table}
\caption{Comparison with attention-augmented ResNet-18 baselines on selected datasets.}
\label{tab:se_cbam}
\centering
\begin{tabular}{lccc}
\toprule
\textbf{Model} & \textbf{Tiny ImageNet} & \textbf{SVHN} & \textbf{FashionMNIST} \\
\midrule
ResNet-18 & 33.02\% & 97.98\% & 64.11\% \\
ResNet-18 + SE & 32.01\% & 97.94\% & 63.11\% \\
ResNet-18 + CBAM & 31.76\% & 97.99\% & 66.98\% \\
\midrule
GradAttn (No PE) & 35.89\% & 98.09\% & 66.70\% \\
GradAttn (Learnable PE) & 34.72\% & 98.09\% & 75.18\% \\
GradAttn (RoPE) & 37.67\% & 98.15\% & 62.40\% \\
\bottomrule
\end{tabular}
\end{table}

The architectural distinction between these approaches is fundamental. SE and CBAM operate as \textit{intra-block} attention mechanisms: They recalibrate feature channels or spatial responses within a single residual block, leaving the gradient pathways between blocks unchanged. GradAttn, by contrast, functions as a \textit{global gradient regulator} across the full network hierarchy. As formalized in Equations~(\ref{eq:meta_regulator}) and~(\ref{eq:gradient_modulation}), the meta-regulator $\mathcal{M}_\psi$ observes a global summary of the network state across all extraction points and emits per-layer modulation signals $\alpha_l \in [0,1]$, governing the effective gradient update $\tilde{g}_l = \alpha_l \cdot \partial\mathcal{L}/\partial\theta_l$ for each stage. This modulation is realised through the transformer self-attention over the token sequence $Z = [z_1, z_2, z_3, z_4, z_5]$ defined in Equation~(\ref{eq:projection}), where the attention operation in Equation~(\ref{eq:attention}) dynamically weights the relative contribution of each hierarchical level. The result is that gradient signals are routed adaptively across the entire network depth rather than recalibrated within any single block. SE and CBAM enhance what is represented within a block; GradAttn controls which blocks drive learning across the network, a fundamentally different and complementary 
form of attention.

The results demonstrate that GradAttn's attention-controlled gradient pathways provide advantages beyond what intra-block channel or spatial attention alone can achieve, validating that inter-layer gradient routing captures complementary information that block-level attention mechanisms cannot access.

\subsection{Discussion}

Domain-Specific Advantages: GradAttn variants excel in domains requiring \textit{global context modeling} (Tiny ImageNet and SVHN) or \textit{abstract structural understanding} (FashionMNIST), but they offer limited benefits in texture-dominated domains.
Complementarity of CNNs and Attention: Attention-based control demonstrates complementary behavior rather than replacing residual connections. Convolutions handle \textit{local feature extraction}, while attention adaptively redistributes gradient signals across scales, enabling dynamic control where deeper features selectively dominate or recede.
Generalization vs. Stability Trade-Off: Gradient diagnostics reveal that minor instability can coincide with improved accuracy, suggesting that \textit{perfectly uniform gradient flow (as in ResNet)} may not be optimal for all tasks, and controlled imbalance through attention may promote richer feature learning.
Comparison with Block-Level Attention: SE and CBAM modules augment individual residual blocks with channel and spatial recalibration, respectively, but they operate within \textit{fixed residual connection topologies}. GradAttn differs fundamentally by routing gradients across the full network hierarchy through \textit{inter-layer attention}, enabling \textit{adaptive weighting of features} from different semantic depths rather than recalibrating features within a single block.

\section{Conclusions} \label{section:conclusion}

This study shows that gradient flow with attention control provides a successful solution to the problem of deep neural networks with fixed residual connections as found in ResNet. The performance of our GradAttn framework was better on five out of eight different datasets and by up to +11.07\% on FashionMNIST when the usual short circuits were replaced by the attention-based pathways that can be modulated by an input and a task to select the most task-relevant feature representations.
The main findings indicate that flow gradient control needs to be adaptive instead of uniform. Through our study of gradient flow, it was uncovered that controlled fluctuations can go hand in hand with enhanced generalization, thus  conflicting with the idea that strict stability is always the best. The variable effectiveness of positional encoding depending on the dataset gives more insights into the architecture; datasets from medical imaging were more compatible with the No PE version that made use of the spatial structure of CNN, whereas intricate natural images had the most significant improvement from the relative positioning of RoPE.
In the future, we will explore larger architectures and gradient-aware training objectives. Treating gradient flow as a learnable component opens pathways for adaptive deep learning architectures beyond residual connections.

\bibliographystyle{unsrt}  
\bibliography{references}

@inproceedings{he2016deep,
  title={Deep residual learning for image recognition},
  author={He, Kaiming and Zhang, Xiangyu and Ren, Shaoqing and Sun, Jian},
  booktitle={Proceedings of the IEEE conference on computer vision and pattern recognition},
  pages={770--778},
  year={2016}
}

@inproceedings{sima2024adaptive,
  title={Adaptive and generic improvements to ResNet Backbone in image classification},
  author={Sima, Zhengdi and Tao, Jingyu and Liu, Zhaochen},
  booktitle={International Conference on Electronics, Electrical and Information Engineering (ICEEIE 2024)},
  volume={13445},
  pages={982--989},
  year={2024},
  organization={SPIE}
}

@article{dosovitskiy2020image,
  title={An image is worth 16x16 words: Transformers for image recognition at scale},
  author={Dosovitskiy, Alexey and Beyer, Lucas and Kolesnikov, Alexander and Weissenborn, Dirk and Zhai, Xiaohua and Unterthiner, Thomas and Dehghani, Mostafa and Minderer, Matthias and Heigold, Georg and Gelly, Sylvain and others},
  journal={arXiv preprint arXiv:2010.11929},
  year={2020}
}

@inproceedings{hu2018squeeze,
  title={Squeeze-and-excitation networks},
  author={Hu, Jie and Shen, Li and Sun, Gang},
  booktitle={Proceedings of the IEEE conference on computer vision and pattern recognition},
  pages={7132--7141},
  year={2018}
}

@inproceedings{huang2017densely,
  title={Densely connected convolutional networks},
  author={Huang, Gao and Liu, Zhuang and Van Der Maaten, Laurens and Weinberger, Kilian Q},
  booktitle={Proceedings of the IEEE conference on computer vision and pattern recognition},
  pages={4700--4708},
  year={2017}
}

@article{jastrzkebski2017residual,
  title={Residual connections encourage iterative inference},
  author={Jastrz{\k{e}}bski, Stanis{\l}aw and Arpit, Devansh and Ballas, Nicolas and Verma, Vikas and Che, Tong and Bengio, Yoshua},
  journal={arXiv preprint arXiv:1710.04773},
  year={2017}
}

@inproceedings{tan2019efficientnet,
  title={Efficientnet: Rethinking model scaling for convolutional neural networks},
  author={Tan, Mingxing and Le, Quoc},
  booktitle={International conference on machine learning},
  pages={6105--6114},
  year={2019},
  organization={PMLR}
}

@inproceedings{woo2018cbam,
  title={Cbam: Convolutional block attention module},
  author={Woo, Sanghyun and Park, Jongchan and Lee, Joon-Young and Kweon, In So},
  booktitle={Proceedings of the European conference on computer vision (ECCV)},
  pages={3--19},
  year={2018}
}

@article{le2015tiny,
  title={Tiny imagenet visual recognition challenge},
  author={Le, Yann and Yang, Xuan},
  journal={CS 231N},
  volume={7},
  number={7},
  pages={3},
  year={2015}
}

@article{krizhevsky2009learning,
  title={Learning multiple layers of features from tiny images},
  author={Krizhevsky, Alex and Hinton, Geoffrey and others},
  year={2009},
  publisher={Toronto, ON, Canada}
}

@inproceedings{netzer2011reading,
  title={Reading digits in natural images with unsupervised feature learning},
  author={Netzer, Yuval and Wang, Tao and Coates, Adam and Bissacco, Alessandro and Wu, Baolin and Ng, Andrew Y.},
  booktitle={NIPS Workshop on Deep Learning and Unsupervised Feature Learning},
  year={2011},
}

@article{xiao2017fashion,
  title={Fashion-mnist: a novel image dataset for benchmarking machine learning algorithms},
  author={Xiao, Han and Rasul, Kashif and Vollgraf, Roland},
  journal={arXiv preprint arXiv:1708.07747},
  year={2017}
}

@article{yang2023medmnist,
  title={Medmnist v2-a large-scale lightweight benchmark for 2d and 3d biomedical image classification},
  author={Yang, Jiancheng and Shi, Rui and Wei, Donglai and Liu, Zequan and Zhao, Lin and Ke, Bilian and Pfister, Hanspeter and Ni, Bingbing},
  journal={Scientific Data},
  volume={10},
  number={1},
  pages={41},
  year={2023},
  publisher={Nature Publishing Group UK London}
}

@inproceedings{veeling2018rotation,
  title={Rotation equivariant CNNs for digital pathology},
  author={Veeling, Bastiaan S and Linmans, Jasper and Winkens, Jim and Cohen, Taco and Welling, Max},
  booktitle={International Conference on Medical image computing and computer-assisted intervention},
  pages={210--218},
  year={2018},
  organization={Springer}
}

@article{pacheco2020pad,
  title={PAD-UFES-20: A skin lesion dataset composed of patient data and clinical images collected from smartphones},
  author={Pacheco, Andre GC and Lima, Gustavo R and Salomao, Amanda S and Krohling, Breno and Biral, Igor P and De Angelo, Gabriel G and Alves Jr, F{\'a}bio CR and Esgario, Jos{\'e} GM and Simora, Alana C and Castro, Pedro BC and others},
  journal={Data in brief},
  volume={32},
  pages={106221},
  year={2020},
  publisher={Elsevier}
}

@inproceedings{xie2017aggregated,
  title={Aggregated residual transformations for deep neural networks},
  author={Xie, Saining and Girshick, Ross and Doll{\'a}r, Piotr and Tu, Zhuowen and He, Kaiming},
  booktitle={Proceedings of the IEEE conference on computer vision and pattern recognition},
  pages={1492--1500},
  year={2017}
}

@inproceedings{szegedy2015going,
  title={Going deeper with convolutions},
  author={Szegedy, Christian and Liu, Wei and Jia, Yangqing and Sermanet, Pierre and Reed, Scott and Anguelov, Dragomir and Erhan, Dumitru and Vanhoucke, Vincent and Rabinovich, Andrew},
  booktitle={Proceedings of the IEEE conference on computer vision and pattern recognition},
  pages={1--9},
  year={2015}
}

@article{zoph2016neural,
  title={Neural architecture search with reinforcement learning},
  author={Zoph, Barret and Le, Quoc V},
  journal={arXiv preprint arXiv:1611.01578},
  year={2016}
}

@inproceedings{ioffe2015batch,
  title={Batch normalization: Accelerating deep network training by reducing internal covariate shift},
  author={Ioffe, Sergey and Szegedy, Christian},
  booktitle={International conference on machine learning},
  pages={448--456},
  year={2015},
  organization={pmlr}
}

@article{ba2016layer,
  title={Layer normalization},
  author={Ba, Jimmy Lei and Kiros, Jamie Ryan and Hinton, Geoffrey E},
  journal={arXiv preprint arXiv:1607.06450},
  year={2016}
}

@inproceedings{wu2018group,
  title={Group normalization},
  author={Wu, Yuxin and He, Kaiming},
  booktitle={Proceedings of the European conference on computer vision (ECCV)},
  pages={3--19},
  year={2018}
}

@inproceedings{wang2018non,
  title={Non-local neural networks},
  author={Wang, Xiaolong and Girshick, Ross and Gupta, Abhinav and He, Kaiming},
  booktitle={Proceedings of the IEEE conference on computer vision and pattern recognition},
  pages={7794--7803},
  year={2018}
}

@article{park2018bam,
  title={Bam: Bottleneck attention module},
  author={Park, Jongchan and Woo, Sanghyun and Lee, Joon-Young and Kweon, In So},
  journal={arXiv preprint arXiv:1807.06514},
  year={2018}
}

@article{dai2021coatnet,
  title={Coatnet: Marrying convolution and attention for all data sizes},
  author={Dai, Zihang and Liu, Hanxiao and Le, Quoc V and Tan, Mingxing},
  journal={Advances in neural information processing systems},
  volume={34},
  pages={3965--3977},
  year={2021}
}

@inproceedings{xiong2020layer,
  title={On layer normalization in the transformer architecture},
  author={Xiong, Ruibin and Yang, Yunchang and He, Di and Zheng, Kai and Zheng, Shuxin and Xing, Chen and Zhang, Huishuai and Lan, Yanyan and Wang, Liwei and Liu, Tieyan},
  booktitle={International conference on machine learning},
  pages={10524--10533},
  year={2020},
  organization={PMLR}
}

@inproceedings{touvron2021training,
  title={Training data-efficient image transformers \& distillation through attention},
  author={Touvron, Hugo and Cord, Matthieu and Douze, Matthijs and Massa, Francisco and Sablayrolles, Alexandre and J{\'e}gou, Herv{\'e}},
  booktitle={International conference on machine learning},
  pages={10347--10357},
  year={2021},
  organization={PMLR}
}

@inproceedings{lin2017feature,
  title={Feature pyramid networks for object detection},
  author={Lin, Tsung-Yi and Doll{\'a}r, Piotr and Girshick, Ross and He, Kaiming and Hariharan, Bharath and Belongie, Serge},
  booktitle={Proceedings of the IEEE conference on computer vision and pattern recognition},
  pages={2117--2125},
  year={2017}
}

@inproceedings{liu2018path,
  title={Path aggregation network for instance segmentation},
  author={Liu, Shu and Qi, Lu and Qin, Haifang and Shi, Jianping and Jia, Jiaya},
  booktitle={Proceedings of the IEEE conference on computer vision and pattern recognition},
  pages={8759--8768},
  year={2018}
}

@inproceedings{ronneberger2015u,
  title={U-net: Convolutional networks for biomedical image segmentation},
  author={Ronneberger, Olaf and Fischer, Philipp and Brox, Thomas},
  booktitle={International Conference on Medical image computing and computer-assisted intervention},
  pages={234--241},
  year={2015},
  organization={Springer}
}

@article{zagoruyko2016wide,
  title={Wide residual networks},
  author={Zagoruyko, Sergey and Komodakis, Nikos},
  journal={arXiv preprint arXiv:1605.07146},
  year={2016}
}

@article{huang2017multi,
  title={Multi-scale dense networks for resource efficient image classification},
  author={Huang, Gao and Chen, Danlu and Li, Tianhong and Wu, Felix and Van Der Maaten, Laurens and Weinberger, Kilian Q},
  journal={arXiv preprint arXiv:1703.09844},
  year={2017}
}

@article{xie2025mhc,
  title={mhc: Manifold-constrained hyper-connections},
  author={Xie, Zhenda and Wei, Yixuan and Cao, Huanqi and Zhao, Chenggang and Deng, Chengqi and Li, Jiashi and Dai, Damai and Gao, Huazuo and Chang, Jiang and Yu, Kuai and others},
  journal={arXiv preprint arXiv:2512.24880},
  year={2025}
}

@inproceedings{wang2023internimage,
  title={Internimage: Exploring large-scale vision foundation models with deformable convolutions},
  author={Wang, Wenhai and Dai, Jifeng and Chen, Zhe and Huang, Zhenhang and Li, Zhiqi and Zhu, Xizhou and Hu, Xiaowei and Lu, Tong and Lu, Lewei and Li, Hongsheng and others},
  booktitle={Proceedings of the IEEE/CVF conference on computer vision and pattern recognition},
  pages={14408--14419},
  year={2023}
}

\end{document}